\documentclass[sigconf]{acmart}

\AtBeginDocument{%
  }

\renewcommand\footnotetextcopyrightpermission[1]{}
\usepackage{times}
\usepackage{soul}
\usepackage{url}
\usepackage{algorithm}
\usepackage{pifont}
\usepackage{xcolor}
\usepackage{algorithmic}
\usepackage[switch]{lineno}
\usepackage{indentfirst}
\usepackage{multirow}
\usepackage{booktabs}
\usepackage{graphicx}
\usepackage{float}
\usepackage{subfig}
\usepackage{xcolor}
\usepackage{newfloat}
\usepackage{listings}

\newtheorem{theorem}{Theorem}
\newtheorem{definition}{Definition}
\newtheorem{proposition}{Proposition}

\author{
Pengyu Zhang, 
Yingjie Liu,
Yingbo Zhou,
Xiao Du,
Xian Wei,
Ting Wang,
Mingsong Chen \\
East China Normal University
}

\begin{document}

\title{When Foresight Pruning Meets Zeroth-Order Optimization: Efficient Federated Learning for Low-Memory Devices}

\begin{abstract}

Although Federated Learning (FL) enables 
 collaborative learning in Artificial Intelligence of Things (AIoT) design, 
it fails to work on low-memory AIoT devices due to its heavy memory usage.
 %
 %
%
To address this problem, various federated pruning methods are proposed to reduce memory usage during inference.  However, few of them can substantially mitigate the memory burdens during pruning and training. As an alternative, zeroth-order or backpropagation-free (BP-Free) methods can partially alleviate the memory consumption, but they suffer from scaling up and large computation overheads,  since the gradient estimation error and floating point operations (FLOPs) increase as the dimensionality of the model parameters grows.
%
In this paper, we propose a federated foresight pruning method based on Neural Tangent Kernel (NTK), which can seamlessly integrate with federated BP-Free training frameworks.
We present an approximation to the computation of federated NTK by using the local NTK matrices. Moreover, we demonstrate that the data-free property of our method can substantially reduce the approximation error in extreme data heterogeneity scenarios.
%
Since our approach improves the performance of the vanilla BP-Free method with fewer FLOPs and truly alleviates memory pressure during training and inference, it makes FL more friendly to low-memory devices. 
Comprehensive experimental results obtained from simulation- and real test-bed-based platforms show that 
our federated foresight-pruning method not only preserves the ability of the dense model with a memory reduction up to $9\times$ but also boosts the performance of the vanilla BP-Free method with dramatically fewer FLOPs.

\end{abstract}

\ccsdesc[500]{Computing methodologies~Distributed artificial intelligence}
\ccsdesc[500]{Computing methodologies~Artificial intelligence}

\keywords{Federated learning, memory efficiency, model pruning, zeroth-order optimization.}

\maketitle

\section{Introduction}

As a promising collaborative learning paradigm in Artificial Intelligence of Things (AIoT) \cite{aiot_survey, date_iot, dac_aiot} design, Federated Learning (FL) \cite{fl_survey} enables knowledge sharing among AIoT devices without compromising their data privacy.  
Specifically, FL enables each device to perform local training and transmit gradients instead of private data. The server then aggregates these local gradients to update the global model~\cite{date_hetero} and redistributes it to a selected subset of devices for the next round of training.
Unlike centralized training deployed on powerful hardware platforms, FL is restricted by stringent local training resource requirements \cite{fl_survey, tcad_hierarchical} reflected by 
being short of adequate computation and memory resources. The expensive communication expenses during the training additionally limit the practical deployment of FL \cite{per_fl_iccad, dac_comm}. 
Moreover, the increasing computation overhead of Deep Neural Networks (DNNs) presents a major barrier to running them on edge devices \cite{sarkar2023edge}. 
Prior work has primarily concentrated on model sparsity methods \cite{pruning_survey, tcad_flcompress} to address these challenges in a one-shot manner. 
By introducing neural network pruning or sparse training approaches \cite{feddst}, the local computation overhead and communication costs are considerably relieved.
Yet, since local devices privately own the data, the standard neural network pruning methods cannot be applied to FL without any modification. The notorious data heterogeneity problem further prohibits local devices from learning a common sparse model \cite{tcad_pervasive}.

\begin{table}[t!]
\centering
\caption{Comparison of peak memory required by FL devices.}
\vspace{-0.05in}
\small
\begin{tabular}{c|ccccc} 
\hline
Method  & \multicolumn{1}{c}{\begin{tabular}[c]{@{}c@{}} Pruning \end{tabular}} & \multicolumn{1}{l}{BP-Free} 
& \multicolumn{1}{c}{\begin{tabular}[c]{@{}c@{}} Peak Device  \\  Memory Cost\end{tabular}} \\ 
\hline
\begin{tabular}[c]{@{}c@{}}FedAvg \cite{mcmahan2017communication} \\\end{tabular}  & \textcolor{red}{\ding{55}}& \textcolor{red}{\ding{55}} 
& $1\textbf{x}$ \\
\begin{tabular}[c]{@{}c@{}}ZeroFL \cite{zerofl}\end{tabular} & \textcolor{green}{\ding{51}} & \textcolor{red}{\ding{55}} 
& $\geq 1\textbf{x}$ \\
\begin{tabular}[c]{@{}c@{}}PruneFL  \cite{prunefl}\end{tabular} & \textcolor{green}{\ding{51}} & \textcolor{red}{\ding{55}}  
& $\geq 1\textbf{x}$ \\
\begin{tabular}[c]{@{}c@{}}FedDST \cite{feddst}\end{tabular} & \textcolor{green}{\ding{51}} & \textcolor{red}{\ding{55}} 
& $\geq 1\textbf{x}$  \\
\begin{tabular}[c]{@{}c@{}}FedTiny \cite{fedtiny}\end{tabular} & \textcolor{green}{\ding{51}} & \textcolor{red}{\ding{55}} 
& $\geq 1\textbf{x}$ \\
\begin{tabular}[c]{@{}c@{}}BAFFLE \cite{does_flzero}\end{tabular} & \textcolor{red}{\ding{55}} & \textcolor{green}{\ding{51}} 
& $<<1\textbf{x}$ \\
\begin{tabular}[c]{@{}c@{}}\textbf{Ours}\end{tabular} & \textcolor{green}{\ding{51}} & \textcolor{green}{\ding{51}}
& $<<1\textbf{x}$\\
\hline
\end{tabular}

\label{compari}
\vspace{-0.15in}
\end{table}

Existing works perform pruning operations under various federated settings \cite{lotteryfl,prunefl,resilient_fl,channel_adapt} to bridge the abovementioned gap.
Although these methods are efficient in inference costs, their training costs regarding memory consumption are seldom investigated.  Since the pruning-during-training methods need to maintain the backpropagation structure \cite{pytorch,tensorflow}, the memory consumption is at least the same as the vanilla training. 
Their heavy dependence on the backpropagation framework prohibits them from truly supporting a lightweight training paradigm for low-memory devices. 
As a result, existing federated pruning methods are not efficient in terms of memory usage;
To eliminate the memory-inefficient auto-differential framework in FL, recent work \cite{does_flzero,zero_1,zero_2}
proposed backpropagation-free (BP-Free) FL methods, aiming to estimate the actual gradients using zeroth-order optimization.
Integrating federated pruning methods and BP-Free FL methods has enormous potential to reduce memory consumption, but it has compatibility gaps and harms performance.
Existing federated pruning steps entangle with gradient computations, making them severely unstable to the precision of the estimated gradient when using a BP-Free strategy.

To mitigate the problems mentioned above, we seek better solutions from the perspective of foresight-federated pruning methods, which disentangle the pruning and training steps. 
Inspired by \cite{does_flzero} and \cite{ntksap}, we propose a federated and BP-Free foresight pruning method to allow lightweight federated pruning. Additionally, we leverage the sparse structure of the pruned model to greatly reduce the local computational overhead and boost the performance of Stein's Identity, a classic zeroth-order optimization method, in FL settings. 
We compared the device peak memory cost of existing outstanding federated learning methods in Table~\ref{compari}. 
Note that our approach explores the synergy between federated pruning and BP-Free training, aiming to 
allow FL on AIoT devices with extremely low memories.  
In summary, this paper makes the following three major contributions:
\begin{itemize}
    \item We propose a novel memory-efficient foresight pruning method for FL, which is resilient to various data heterogeneities.
    \item We propose an approximation to the federated NTK matrix and show that the data-free property of our method can effectively reduce the approximation error.
    \item We implement our proposed foresight pruning method in conjunction with BP-Free training and conduct comprehensive experiments on both simulation- and real test-bed-based platforms to demonstrate the effectiveness of our approach.
\end{itemize}

\section{Related Work}
\noindent\textbf{Federated Neural Network Pruning.}
A two-step pruning method was proposed in PruneFL \cite{prunefl} to reduce the computation cost on the device side.
It first completes coarse pruning on a selected device. Further, finer pruning is conducted on the server side to reduce computation expenses. 
ZeroFL \cite{zerofl} separates model parameters into
active and non-active groups for the purpose of the efficiency of both forward and backward passes. 
Although the aforementioned methods leave the finer pruning on the server side, they still require a large local memory footprint to record the updated importance scores of all parameters for the guidance of pruning. 
FedDST \cite{feddst}, instead, adopts sparse training, which consists of mask adjustment and regrow on the device side. Once the local sparse training is finished, the server conducts sparse aggregation and magnitude pruning procedures to obtain a global model. 
Even with the sparse training framework, the high requirements on memory are not relieved. 
Worse still, the local mask adjustment and regrow processes inevitably need additional training epochs, which contradicts the principle of minimizing local training steps in FL \cite{localstep}.
To reduce the cost of the memory- and computation-intensive pruning process for extremely recourse-constrained devices, FedTiny \cite{fedtiny} adaptively searches finer-pruned specialized models by progressive pruning based on a coarsely pruned model. Effective as it is under the low-density region for inference, the training step is hindered by the significant communication and memory burdens, making it less compatible with devices with limited memory capacity.

\noindent\textbf{BP-Free Federated Learning.}
Gradient-based optimization techniques are commonly used for training DNNs in FL settings.
Regardless, the backpropagation (BP) operations heavily depend on the auto-differential framework (e.g., Pytorch \cite{pytorch}). The additional static and dynamic memory requirements are unaffordable to low-memory devices.
Recently, advances in zeroth-order optimization methods \cite{zero_1,zero_2,zero_trajectory} have demonstrated potential for federated training, particularly in cases where backward processes are expensive from the perspective of computation and memory space. \cite{does_flzero} proposes to utilize Stein's Identity \cite{stein} as the gradient estimation method to develop a BP-Free federated learning framework. Yet, the Floating Point Operations (FLOPs) per communication round are significantly increased to an extremely high level during the training due to the property of BP-Free gradient estimation: numerous forward passes are required for one-time gradient estimation to decrease the estimation error \cite{sharp_covariances,does_flzero}. The degradation of learning performance is another issue since the BP-Free training comes at the cost of adding noises \cite{noise_1,noise_2} to the gradients. Therefore, we propose using sparsity techniques to reduce the FLOPs and boost learning performance in a one-shot manner.

\section{Preliminaries}
Let $\mathcal{D}=\{\mathcal{X}, \mathcal{Y}\}$ denote the entire training dataset. 
Let $N$ represent the total number of data points.
Hence, we have
$\mathcal{X}=\{\mathbf{x}_1, \cdots, \mathbf{x}_N\}$ and $\mathcal{Y}=\{\mathbf{y}_1, \cdots, \mathbf{y}_N\}$ denote inputs and labels, respectively. 

\noindent\textbf{Federated Learning.}
FL aims to collaboratively learn a global model parameterized by $\boldsymbol{W}\in\mathbb{R}^{n}$ while keeping local data private. 
%
Given that $m$ devices are involved in each round of local training, where the local dataset on $i$-th device is denoted as 
$\mathcal{D}_i={(\mathbf{x}_{j}, \mathbf{y}_{j})}_{j=1}^{N_j}$ with
$N_j$ representing the number of data points, the objective is defined by
\begin{align} 
\label{fedavg}
    \mathop{\min}\limits_{\boldsymbol{W}} f(\boldsymbol{W})= \sum\limits_{i=1}^{m}\frac{1}{N_i}\sum\limits_{j=1}^{N_j}\mathcal{L}(\boldsymbol{W};(\mathbf{x}_{j},\mathbf{y}_{j})),
\end{align}
where $\mathcal{L}(\boldsymbol{W}; \mathcal{D}_i)$ is the specified loss function on local dataset $\mathcal{D}_i$.


\noindent\textbf{Neural Tangent Kernel.}
Neural Tangent Kernel (NTK) analyzes the training dynamics of neural networks \cite{ntk_jacot}. 
Given an arbitrary DNN $f$ initialized by $\boldsymbol{W}_0$, the NTK at initial state is defined as 
\begin{align*}
    \boldsymbol{\theta}_0 =\langle\nabla_{\boldsymbol{W}_0}f(\mathcal{X};\boldsymbol{W}_0), \nabla_{\boldsymbol{W}_0}f(\mathcal{X};\boldsymbol{W}_0)\rangle.
\end{align*}


It has been proven that the NTK stays asymptotically constant during the training if the DNN is sufficiently large in terms of parameters. Therefore, the NTK at initialization (i.e., $\boldsymbol{\theta}_0$) can characterize the training dynamics.

\noindent\textbf{Foresight Pruning Based on NTK.}
The general objective function of foresight pruning is formulated as
\begin{align}\label{gen_obj}
\small
\min _{\mathbf{m}} \mathcal{L}(\mathcal{A}(\boldsymbol{W}_{0}, \mathbf{m}); \mathcal{D}) \;
\text{s.t.} \; \mathbf{m} \in\{0,1\}^{p}, \;\|\mathbf{m}\|_{0} / n \leq d, 
\end{align}
where $\mathcal{A}$ denotes the model architecture dominated by the binary mask $\mathbf{m}$, $\boldsymbol{W}_0$ represents the model parameter at initialization. We aim to find the mask $\mathbf{m}$ in which each element follows the binary distribution that minimizes the loss function and is simultaneously constrained by the target density $d$. 
To make Eq.~\ref{gen_obj} tractable,
existing foresight pruning methods propose the saliency measurement function, defined as
\begin{align} \label{saliency_score}
S(\mathbf{m}^{j})=S(\boldsymbol{W}_{0}^{j})=\frac{\partial \mathcal{I}}{\partial \mathbf{m}^{j}}=\frac{\partial \mathcal{I}}{\partial \boldsymbol{W}_{0}^{j}} \cdot \boldsymbol{W}_{0}^{j},
\end{align}
where $\mathcal{I}$ represents a function of model parameters and mask $\mathbf{m}$.
$\mathbf{m}^{j}$ denotes the $j$-th element of the mask $\mathbf{m}$, which is a scalar.
$S(\mathbf{m}^{j})$ represents the saliency score function that measures the impact of deactivating the $\mathbf{m}^{j}$ (i.e., set $\mathbf{m}^{j}$ to $0$). After the saliency score is computed, we keep the top-$d$ elements in the mask. In detail, we set the element in the mask $1$ ($\mathbf{m}^{j}=1$) if it belongs to the set of top-$d$ elements and $0$ ($\mathbf{m}^{j}=0$) otherwise. To analyze the property of $\boldsymbol{\theta}_0$, we follow the work proposed by \cite{ntksap} to compute the spectrum of it.
The spectrum is formulated as
\begin{align} \label{ntk_I}
   ||\boldsymbol{\theta}_0||_{*} = ||\boldsymbol{\theta}_0||_{\mathbf{tr}} = ||\nabla_{\boldsymbol{W}_0}f(\mathcal{X};\boldsymbol{W}_0)||_F^2,
\end{align}
where $||\cdot||_{*}$ is the nuclear norm operation. The trace norm $||\boldsymbol{\theta}_0||_{\mathbf{tr}}$ is equivalent to the nuclear norm since the NTK matrix is symmetric. To further reduce the cost of computing the NTK matrix, we rewrite the trace norm of the NTK matrix as the Frobenius norm of gradients $||\nabla_{\boldsymbol{W}_0}f(\mathcal{X};\boldsymbol{W}_0)||_F^2$. We aim to find a sparse model with the same training dynamics as the dense model. Therefore, the objective function of NTK-based foresight pruning is formulated as
\begin{align}\label{saliency function}
S(\mathbf{m}^{j})=\left|\frac{\partial||\nabla_{\boldsymbol{W}_0}f(\mathcal{X};\boldsymbol{W}_0\odot\mathbf{m})||_F^2}{\partial \mathbf{m}^{j}}\right|.
\end{align}
Note that to achieve a BP-Free method, the Fobenius norm of gradients can be
approximated by 
\begin{align}\label{norm_fd}
    \left\|f(\mathcal{X}; \boldsymbol{W}_{0} \odot \mathbf{m})-f(\mathcal{X}
;(\boldsymbol{W}_{0}+\Delta \boldsymbol{W}) \odot \mathbf{m})\right\|_{2}^{2},
\end{align}
which saves the expensive computation of Eq.~\ref{saliency function}.
%

\noindent\textbf{Finite Difference and Stein's Identity.}
Derived from the definition of derivatives, the Finite Difference (FD) method can be extended to multivariate cases 
 by using Taylor's expansion.
Given the loss function $\mathcal{L}(\boldsymbol{W};\mathcal{D})$ and a small perturbation $\boldsymbol{\delta}\in\mathbb{R}^{n}$, the FD method is define as
\begin{align}
    \mathcal{L}(\boldsymbol{W}+\boldsymbol{\delta};\mathcal{D}) - \mathcal{L}(\boldsymbol{W};\mathcal{D}) = \boldsymbol{\delta}^T\nabla_{\boldsymbol{W}}\mathcal{L}(\boldsymbol{W}) + o(||\boldsymbol{W}||_2^2).
\end{align}
Assuming the loss function is continuously differentiable w.r.t. model parameter $\boldsymbol{W}$. The precise gradient of $\boldsymbol{W}$ estimated by Stein's Identity is formulated as 
\begin{equation}
\label{stein}    \nabla_{\boldsymbol{W}}\mathcal{L}(\boldsymbol{W}) = \mathbb{E}_{\boldsymbol{\delta}\sim\mathcal{N}(0,\sigma^2\mathbf{I})}[\frac{\boldsymbol{\delta}}{\sigma^2}\Delta\mathcal{L}(\boldsymbol{W};\mathcal{D})],
\end{equation}
where $\Delta\mathcal{L}(\boldsymbol{W};\mathcal{D}) = \Delta\mathcal{L}(\boldsymbol{W}+\boldsymbol{\delta};\mathcal{D}) - \Delta\mathcal{L}(\boldsymbol{W};\mathcal{D})$,  $\boldsymbol{\delta}$ follows a Gaussian distribution with zero mean and covariance $\sigma^2\mathbf{I}$. We  utilize the Monte Carlo method to sample $K$ number of $\boldsymbol{\delta}$, thereby obtaining a stochastic version of the estimation as
\begin{align}
\label{monte_stein}
\widehat{\nabla_{\boldsymbol{W}}}\mathcal{L}(\boldsymbol{W}) = \frac{1}{K}\sum_{k=1}^{K}[\frac{\boldsymbol{\delta}_k}{\sigma^2}\Delta\mathcal{L}(\boldsymbol{W};\mathcal{D})].
\end{align}

\section{Methodology}

\subsection{Estimation Error of Stein's Identity}
Though the sampling method in Eq.~\ref{monte_stein} is friendly to low-memory devices, the computation overhead is extremely heavy since $K$ should be large for accurate estimation.
Following what was proven in \cite{does_flzero}, the estimation error is formally defined by the following theorem:
\begin{theorem}
(Estimation error \cite{does_flzero}) Let  $\hat{\boldsymbol{\delta}}=\frac{1}{K}\sum_{k=1}^K\boldsymbol{\delta}_k$, covariance matrix $\widehat{\boldsymbol{\Sigma}}=\frac{1}{K\sigma^2}\sum_{k=1}^K\boldsymbol{\delta}_k\boldsymbol{\delta}_k^T$. Let $n$ be the dimension of trainable parameters $\mathbf{W}$ of the DNN. The discrepancy between the true gradient $ \nabla_{\boldsymbol{W}} \mathcal{L}(\mathbf{W} ; \mathbf{D})$ and the estimated gradient $\widehat{\nabla_{\boldsymbol{W}}} \mathcal{L}(\mathbf{W} ; \mathbf{D})$ is formulated as
\begin{align*}
    \widehat{\nabla_{\boldsymbol{W}}} \mathcal{L}(\mathbf{W} ; \mathbf{D})=\widehat{\boldsymbol{\Sigma}} \nabla_{\boldsymbol{W}} \mathcal{L}(\mathbf{W} ; \mathbf{D})+o(\widehat{\boldsymbol{\delta}})&; \\
    \text { s.t. } \mathbb{E}[\widehat{\boldsymbol{\Sigma}}]=\mathbf{I}, \mathbb{E}[\widehat{\boldsymbol{\delta}}]=\mathbf{0}&.
\end{align*}
\end{theorem}

Neglecting the trivial term $o(\widehat{\boldsymbol{\delta}})$, the estimation error measured by Mean Square Error (MSE) is fully controlled by the term $||\hat{\boldsymbol{\Sigma}}-\mathbf{I}||_2^2$. Based on the proof in \cite{sharp_covariances}, we have $||\hat{\boldsymbol{\Sigma}}-\mathbf{I}||_2 \leq\sqrt{\frac{n}{K}}$. In conclusion, we can expect a more accurate estimation if we either increase the number of Monte Carlo steps or decrease the dimensionality of the model parameter $\boldsymbol{W}$, or both. 
In practice, $K$ is related to the total FLOPs consumed to estimate the true gradient, and $n$ reflects the memory requirements for a device to afford the model. 
Developing more advanced zeroth-order algorithms to lower the value of $K$ required to achieve the same performance is a promising way to obtain more accurate estimations. Another way is to put effort into decreasing the value of $n$.
To achieve the latter, we introduce model pruning to reduce computation overhead and increase the estimation precision as it aims to find the spare structure of the model, and can still maintain the performance. 

\begin{algorithm}[h]
\small
    \caption{Pruning and BP-Free Training}
    \label{alg}
\begin{flushleft}
\textbf{Input}: 
1) $\boldsymbol{W}_0$, a randomly initialized global model; 
2) $C$, a pool of all participants; 
3) $T_p$, \# of pruning rounds; 
4) $T_t$, \# of training rounds; 
5) $G_p$, \# of participants for pruning; 
6) $G_t$, \# of participants for training; 
7) $K$, \# of perturbations to estimate gradients; 
8) $d$, target density; 
9) $\mathbf{m}$, mask for parameters; 
10) $\eta$, learning rate.
\\
\textbf{NTK Foresight Pruning:}
\end{flushleft}
\begin{algorithmic}[1] 
\FOR{$t=1,\dots,T_p$}\label{line:trainStart}
    \STATE $C_t$ $\leftarrow$ $\text{Random\ Sample}(C,G_p)$; \\
    \STATE $G_p$ $\leftarrow$ Number of elements in $C_t$; \\
    \IF{use real data}
        \FOR {$ i = 1,\dots, G_p$ in parallel}
            \STATE Get $F_i$ based on Eq.~\ref{norm_fd_fl} at devices;
            \STATE Send $F_i$ to the server;
        \ENDFOR
    \ELSIF{use random data}
    \STATE Sample $\mathbf{x}_{i}$ from the standard Gaussian Distribution;
    \STATE Get $F_i$ based on Eq.~\ref{norm_fd_fl} at the server;
    \ENDIF
    \STATE Get $\mathcal{I}$ based on Eq.~\ref{I_fl};
    \STATE Compute $\mathcal{S}(\mathbf{m}^j)$ based on Eq.~\ref{saliency_fl} ;
    \STATE Get threshold $\tau$ as $(1-d^{\frac{t}{T_p}})$ percentile of $\mathcal{S}(\mathbf{m}^j)$;
    \STATE $\mathbf{m}$ as $\mathbf{m}\leftarrow\mathbf{m}\odot\mathcal{S}(\mathbf{m}^j)<\tau$;
\ENDFOR
\STATE \textbf{Return} A pruned model parameterized by $\boldsymbol{W}_0\odot\mathbf{m}$
\end{algorithmic}
\begin{flushleft}
    \textbf{BP-Free Training:}
\end{flushleft}
\begin{algorithmic}[1] 
\STATE $\boldsymbol{W} \leftarrow \boldsymbol{W}_0\odot\mathbf{m}$;
\FOR{$t=1,\dots,T_t$}\label{line:trainStart2}
    \STATE $C_t$ $\leftarrow$ $\text{Random\ Sample}(C,G_t)$; \\
    \STATE $G_t$ $\leftarrow$ Number of elements in $C_t$; \\
    \FOR {$ i = 1,\dots, G_t$ in parallel}
        \STATE Set $\boldsymbol{W}_i \leftarrow \boldsymbol{W}$; \\
        \STATE Get $\widehat{\nabla_{\boldsymbol{W}_i}}\mathcal{L}(\boldsymbol{W}_i)$ based on Eq.~\ref{monte_stein};
    \ENDFOR
    \STATE Devices send $\Delta\mathcal{L}(\boldsymbol{W}_i;\mathcal{D}_i)$ and corresponding random seed to the server;
    \STATE The server produces $\boldsymbol{\delta}$ after receiving random seed;
    \STATE Get $\widehat{\nabla_{\boldsymbol{W}_{agg}}}\mathcal{L}(\boldsymbol{W})$ based on Eq.~\ref{agg_grad};
    \STATE $\boldsymbol{W}\leftarrow\boldsymbol{W}-\eta\widehat{\nabla_{\boldsymbol{W}_{agg}}}\mathcal{L}(\boldsymbol{W})$;
\ENDFOR
\STATE \textbf{Return} A global model parameterized by $\boldsymbol{W}$
\end{algorithmic}
\end{algorithm}

Current federated pruning methods are highly integrated with the training process. Consequently, the noisy gradients negatively affect the pruning step, decreasing training performance.
To mitigate the impact of noisy gradients on the pruning step, we propose decoupling the pruning processes from the training processes, which is one of the fundamental motivations of foresight pruning. Further, we dedicate efforts to developing a federated foresight-pruning method based on the NTK spectrum, ensuring full compatibility with the BP-Free training framework.


\subsection{NTK-based Foresight Pruning for FL}
Since the symmetric property of the $\boldsymbol{\theta}_0$ is only valid for centralized training, we cannot directly apply it for federated training. Motivated by \cite{huang2021flntk}, we prove that the NTK in federated learning can be approximated by local NTK matrices.
\begin{definition} \label{fl-ntk}
(Asymmetric FL-NTK) 
Let $\boldsymbol{\theta}_0^i \in\mathbb{R}^{N_i \times N_i}$ represent the local NTK of the $i^{th}$ device based on its dataset. Let $N_{max}$ represent $\max\{N_1, N_2,..., N_m\}$ and $N_{sum}$ represent $\sum_{i=1}^{m} N_i$. The asymmetric FL-NTK is defined as 
$\boldsymbol{\theta}_0^{fl}\in\mathbb{R}^{N_{max} \times N_{sum}}$.
The $\boldsymbol{\theta}_0^{fl}$ is formulated by combining the $N_i$ columns of $\boldsymbol{\theta}_0^i$ for all $i \in m$ and padding the reset elements with zero.
\end{definition}
The asymmetric property of FL-NTK prevents us from computing the nuclear norm by the Frobenius norm of gradients based on each local dataset. 
However, we prove that we can utilize the summation of local symmetric NTK matrices to approximate the FL-NTK and efficiently conduct the Frobenius norm computation.

\begin{proposition}
Since $\boldsymbol{\theta}_0^{fl}$ is the horizontal concatenation of $\boldsymbol{\theta}_0^{i}$ for $i\in m$, we decompose it as the summation of $m$ sparse NTK matrices. Let $\boldsymbol{\theta}_{0,sp}^i \in\mathbb{R}^{N_{max} \times N_{sum}}$ represent the sparse NTK matrix for the $u^{th}$ device. We can reformulate the FL-NTK matrix as
\begin{align}
&\boldsymbol{\theta}_0^{fl}=\sum_{i=1}^{m}\boldsymbol{\theta}_{0,sp}^i, \\[-1mm]
\boldsymbol{\theta}_{0,sp}^i[0:N_i, \sum_{j=1}^{i}&N_j:\sum_{j=1}^{i}N_j+N_i] = \boldsymbol{\theta}_0^{i},
\end{align}
where the non-filling elements are all zeros in $\boldsymbol{\theta}_{0,sp}^i$.
\end{proposition}
Since the nuclear norm satisfies the triangle inequality, we further formulate the computation of $\|\boldsymbol{\theta}_0^{fl}\|_{*}$ as
\begin{align}
\label{flntk}
    \centering
    \hspace{-4mm}
    \|\boldsymbol{\theta}_0^{fl}\|_{*}= \|\sum_{i=1}^{m}\boldsymbol{\theta}_{0,sp}^i\|_{*}\leq\sum_{i=1}^{m}\|\boldsymbol{\theta}_{0,sp}^i\|_{*}   =\sum_{i=1}^{m}\|\boldsymbol{\theta}_{0}^i\|_{*}
\end{align}
Therefore, the complex FL-NTK is upper-bounded by the summation of individual local NTK matrices. Since the sparse model preserves the same training dynamic as the dense model and estimated gradients based on Stein's Identity are unbiased, the overall convergence rate based on the training round $T$ is $O(1/\sqrt{T})$, the same as the standard FedAvg \cite{li2023convergence} algorithm under non-convex settings.

Unlike centralized pruning, the entire dataset $\mathcal{D}$ is distributed to multiple devices in FL. 
According to the form of Eq.~\ref{saliency_score}, the computation of saliency score demands at least one operation of gradient computing, i.e., $\partial\mathcal{I} / \partial\boldsymbol{W}_0^j$ if not consider the function $\mathcal{I}$. We can wisely let the local devices compute the function $\mathcal{I}$ and leave the derivative and multiplication operations in Eq.~\ref{I_fl} to the server to complete. In this way, local datasets equally contribute to the generation of the binary mask, yielding an unbiased global sparse model, and the memory required to support the auto-differential framework is shifted to the server. Things might be different when considering the specific form of the function $\mathcal{I}$. 
As mentioned above, the function $\mathcal{I}$ defined in Eq.~\ref{ntk_I} can be approximated by 
either 
Eq.~\ref{norm_fd} or 
the loss gradient w.r.t. weights
$\left\|\nabla_{\boldsymbol{W}_{0}} \mathcal{L}\right\|_{2}^2$. 
The latter introduces extra gradient computation, which inevitably relies on the auto-differential framework.
In contrast, Eq.~\ref{norm_fd} benefits from being independent of the auto-differential framework. 
Regardless, the computation of the saliency score in Eq.~\ref{saliency_score} still intertwines with it (i.e., $\partial\mathcal{I} / \partial\boldsymbol{W}_0^j$). We might introduce another zeroth-order optimization method to estimate the derivatives, but it introduces more estimation errors. 
As a result, we choose Eq.~\ref{norm_fd} as the function $\mathcal{I}$ to make the local devices free of the memory-inefficient backpropagation framework.
The federated NTK foresight pruning is organized as
\begin{align} 
\hspace{-3mm}
F_i = \|f(\mathbf{x}_{i} ; \boldsymbol{W}_{0, i} \odot \mathbf{m}&)-f(\mathbf{x}_{i} ;(\boldsymbol{W}_{0, i}+\Delta \boldsymbol{W}_i ) \odot \mathbf{m})\|_{2}^{2},
\label{norm_fd_fl}
\\
\footnotesize
S(\mathbf{m}^{j})&=\left|\frac{\partial\mathcal{I}}{\partial \mathbf{m}^{j}}\right| = \left|\frac{\partial \mathcal{I}}{\partial \boldsymbol{W}_{0}^{j}} \cdot \boldsymbol{W}_{0}^{j}\right|, 
\label{saliency_fl}
\\
\mathcal{I} = \frac{1}{N} &\sum\nolimits_{i=1}^{N} \mathbb{E}_{\Delta \boldsymbol{W} \sim \mathcal{N}(\mathbf{0},\epsilon \mathbf{I})}[F_i], 
\label{I_fl} 
\end{align}
where $i$ represents the $i$-th dataset owned by the corresponding device. 
$\boldsymbol{W}_0^j=(\boldsymbol{W}_{0,i}^j)_{i=1}^{m}$ denotes the $j$-th element of the initial global model.
$\Delta\boldsymbol{W}_i$ is sampled from a Gaussian distribution with zero mean and $\epsilon\mathbf{I}$ variance. 
The $\mathbb{E}$ symbol denotes the expectation over $\Delta\boldsymbol{W}_i$, which can be approximated by the Monte Carlo sampling method. To avoid layer-collapse, we use an exponential decay schedule $\tau=(1-d^{t/T_p})$ to compute the pruning threshold $\tau$, where $t$ denotes the current pruning round, $d$ is the target density and $T_p$ is the maximum pruning round. The parameters whose saliency scores are under the threshold $\tau$ will be pruned.
It may be noted that the proposed NTK pruning method introduces $T_p$ extra computations to the devices. In addition, the multiple-round pruning schedule requires communication between the server and all devices. 
To resolve the two problems in one shot, we fully explore the intrinsic property of NTK pruning. 
Motivated by \cite{nasi}, we show that under data heterogeneity scenarios, the difference between any twp local NTK matrices is $\Big|\left\|\boldsymbol{\theta}_0^i(P^i)\right\|_{*}-\left\|\boldsymbol{\theta}_0^j(P^j)\right\|_{*}\Big|\leq n_0^{-1}Z$, where $P^i$ and $P^j$ represent the local data distributions, $Z$ represent $\int\left\|P^i(\mathbf{X})-P^j(\mathbf{X})\right\|\mathrm{d}\mathbf{X}$. $n_0$ is the input dimension. In our cases, $n_0$ is $1024$ and $0<Z<2$. Therefore, the approximation error defined by the summation operation in Eq.~\ref{flntk} can be represented by the difference and increases as the degree of heterogeneity grows. To alleviate the approximation error, we constraint the difference between $P^i$ and $P^j$ by sampling the data $\mathbf{x}_{i}$ in Eq.~\ref{norm_fd_fl} from the same standard Gaussian distribution, which diminishes the difference between two NTK matrices and makes our method resilient to data heterogeneity.
Moreover, the computation of Eq.~\ref{norm_fd_fl} can be fully conducted by the server, and our proposed federated NTK pruning method degrades to the centralized version, entirely saving the communication overhead.

\subsection{Backpropgation-free Training for FL}
We further integrate the proposed NTK foresight pruning with Stein's Identity method to collaboratively build a memory-friendly federated training framework. To leverage Stein's Identity in the context of FL, we rewrite the estimation of the aggregated gradient in the form of FedAvg \cite{fedavg} as
\begin{align} \label{stein_avg}
\widehat{\nabla_{\boldsymbol{W_{agg}}}}\mathcal{L}(\boldsymbol{W}) = \sum_{i=1}^{m}\frac{N_i}{KN}\sum_{k=1}^{K}[\frac{\boldsymbol{\delta}}{\sigma^2}\Delta\mathcal{L}(\boldsymbol{W}_i;\mathcal{D}_i)].
\end{align}
Note that $\boldsymbol{\delta}\in\mathbb{R}^{n}$ and $n$ is the dimensionality of gradients we want to estimate. 
%
%
To further decrease the communication burden, 
we use the \textit{Random Seed Trick} technique. 
The \textit{Random Seed Trick} smartly leverages the inherent property of the developing environment: i) $\boldsymbol{\delta}$ has the same dimensionality as the gradients we want to estimate and is generated by sampling from a Gaussian distribution; ii) the sampling from Gaussian distribution is driven by the \textit{Random Seed} in software level. Therefore, the server can produce the same sampling results as the local devices by maintaining the same \textit{Random Seed} and running environments between them. In detail, Eq.~\ref{stein_avg} is performed by the server and devices as:
\begin{align} \label{agg_grad}
\footnotesize
     \widehat{\nabla_{\boldsymbol{W}_{agg}}}\mathcal{L}(\boldsymbol{W}) 
     = \sum_{i=1}^{m}\frac{N_i}{KN}\sum_{k=1}^{K}[\frac{\overbrace{\boldsymbol{\delta}}^{\text{server produces}}}{\sigma^2}\underbrace{\Delta\mathcal{L}(\boldsymbol{W}_i;\mathcal{D}_i)}_{\text{computed by devices}}].
\end{align}
In practice, we let the devices compute $\Delta\mathcal{L}(\boldsymbol{W}_i;\mathcal{D}_i)$ and leave the $\boldsymbol{\delta}$ to the server.
Consequently, the communication cost is significantly reduced from large dimensional vectors to scalars level since we do not need to transmit the vector $\boldsymbol{\delta}$. The overall training processes are shown in Algorithm~\ref{alg}.




\section{Experiments}

In this section, we conducted federated pruning and federated zeroth-order experiments to answer the following two pivotal \textbf{R}esearch \textbf{Q}uestions:
\textbf{RQ1}: What are the advantages of our NTK-based federated foresight pruning? \textbf{RQ2}:  What improvements do the NTK-based foresight federated pruning bring to the standard federated zeroth-order method?


\vspace{-2mm}
\subsection{Experimental Settings}
\noindent\textbf{Dataset.}
We evaluated the performances on two image classification tasks, i.e., 
CIFAR-10, CIFAR-100 \cite{cifar}.
Following the heterogeneity configurations in \cite{silos}, we considered Dirichlet distribution throughout this paper, which is parameterized by a coefficient $\beta$, denoted as $Dir(\beta)$. $\beta$ determines the degree of data heterogeneity. The smaller the $\beta$ is, the more heterogeneous the data will be. 
We only considered one non-IID scenario by setting $\beta$ as  $0.1$.

\noindent\textbf{Models.}
To show the scalability of proposed foresight pruning to various types of models, we utilized the LeNet-5 \cite{Lenet} and ResNet-20 \cite{residual} with batch normalization.
For federated pruning evaluation experiments, we utilized LeNet-5 for CIFAR-10 and ResNet-20 for CIFAR-100.
For the experiments on BP-Free federated training, we used LeNet-5 and ResNet-20 for CIFAR-10 and CIFAR100 datasets.

\noindent\textbf{Baselines.}
For the pruning evaluation, we included FedDST and PruneFL to 
compare with the proposed NTK pruning. FedDST utilizes a sparse training method and conducts neuron growth and pruning at local devices. The 
PruneFL first conducts local coarse pruning at a randomly selected device, then performs server-side pruning based on collected gradients from local devices.
We excluded the federated pruning methods that are inapplicable for memory-constrained FL (i.e., ZeroFL, LotteryFL). 
To show the improvements brought by our NTK pruning to BP-Free federated learning, we compared it with the state-of-the-art method BAFFLE \cite{does_flzero}, which conducted the vanilla Stein's Identity method.
Note that the result of FedAvg is conducted by backpropagation-based training.

\noindent\textbf{Hyperparameter Settings.}
The pruning rounds $T_p$ for LeNet and ResNet-20 are set to $50$ and $20$, respectively.
The corresponding learning rate and batch size are set to $0.001$ and $256$.
Following the source code in \cite{feddst}, we set the number of local training epochs to $5$ if not specified.
We randomly selected $10$ out of $100$ devices to perform local training at each round, and the training batch size is set to $32$.
For experiments on CIFAR-10, we set the learning rate, momentum, and weight decay as $0.01$, $0.9$, and $1e-3$, respectively. For the CIFAR-100, the momentum and weight decay are set to $0$, $1e-3$ for both models. Specifically, we set the learning rate to $0.01$ for LeNet and $0.1$ for ResNet. The learning rate decays to $99.8\%$ after each round for all experiments.
For all experiments, we set $K$, the number of Monte Carlo steps to estimate the true gradient, to $50$, $100$, and $200$, respectively. For a fair comparison, we used a sparsity level of $80$\% and $5$ epoch training based on the original FedDST paper.

\subsection{Experimental Results for Federated Pruning}
\begin{figure}[h]  
    \centering
    \vspace{-0.2in}
\includegraphics[width=1\linewidth]
    {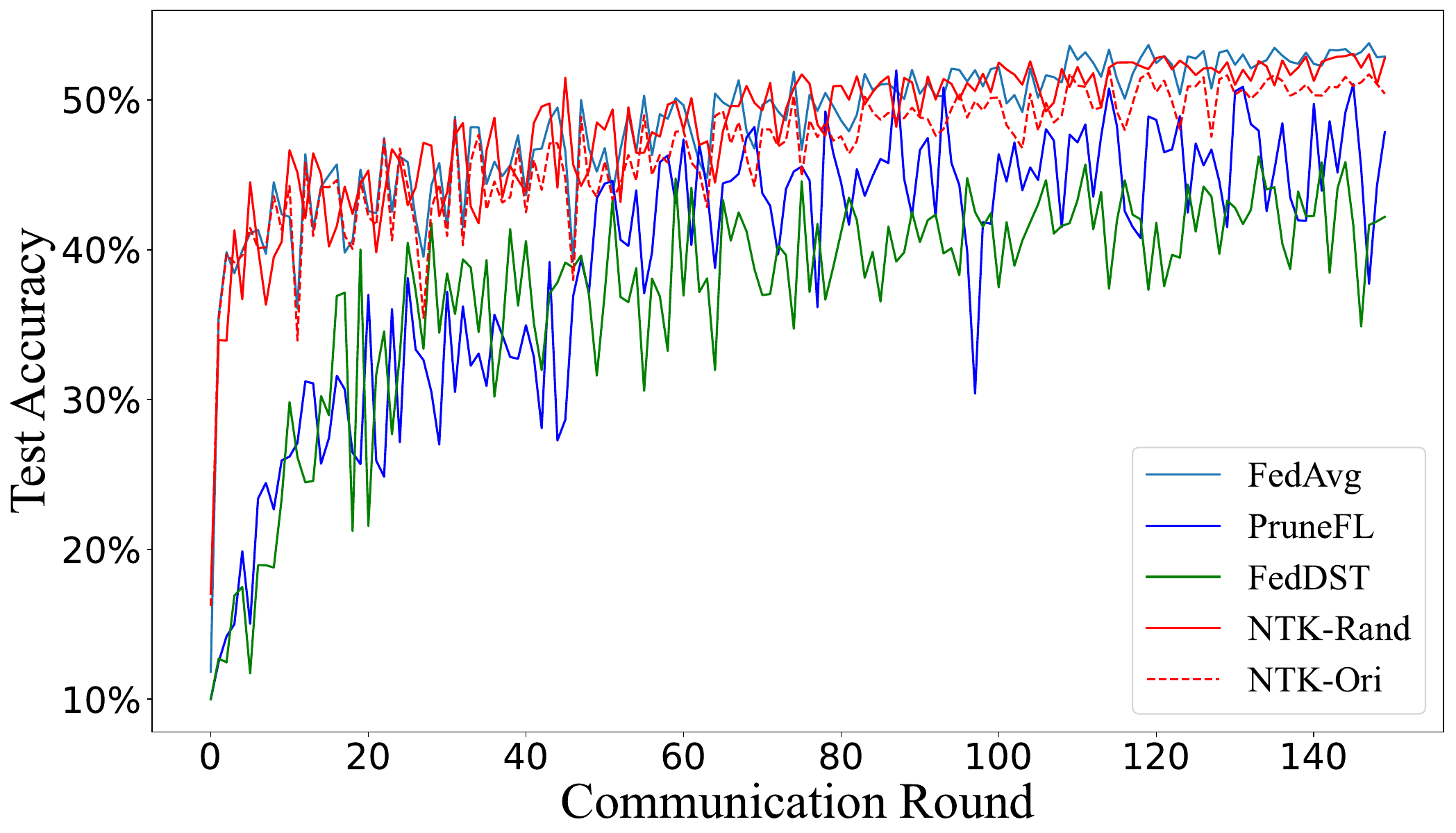}
    \vspace{-0.25in}
    \caption{Test accuracy comparison on CIFAR-10.}
      \label{cifar-10-non-fw}
\end{figure}

\noindent\textbf{Accuracy and FLOPs Analysis.}
Table~\ref{acc-round} shows the maximum accuracy and FLOPs for all federated pruning baselines under an extreme non-IID scenario where $\beta=0.1$. The training curves for CIFAR-10 are shown in Figure~\ref{cifar-10-non-fw}. Note that ``NTK-Ori'' represents the result on real data and ``NTK-Rand'' represents that on randomly generated data. We denote epoch as ``ep'' in tables. Since the time coefficient is required to implement PruneFL, we omit PruneFL for the ResNet-20 model as the source code of FedDST or PruneFL does not support the corresponding time coefficient.
The sparsity of all experiments is initialized to $80\%$.
We use the FLOPs to evaluate whether local devices suffer
from intensive computation overhead, thus showing the efficiency of the federated learning algorithm.

For the Fedavg method, the accuracy after convergence is $53.66\%$. Since it is conducted on a dense model, the FLOPs are not reduced, making it inefficient for resource-constrained devices. For the FedDST method, the FLOPs are significantly reduced to $20\%$ of the dense model. However, the accuracy drop is not negligible. In addition, the convergence speed is slower than the FedAvg method due to the consistent neuron growth and pruning steps. For the PruneFL approach, the accuracy drop after convergence is less than that of FedDST. Note that PruneFL in our experiment ends up with an all-one mask, which makes the model dense again when convergences. Therefore, the FLOPs of PruneFL remain the same compared to FedAvg. 
Since our NTK method collects pruning information from all local devices, the accuracy drop is largely reduced to $0.42\%$. Moreover, the NTK method avoids model sparsity readjustment during the training. As a result, the training stability shown in Figure~\ref{cifar-10-non-fw} is consolidated compared with FedDST and PruneFL.
Though the FLOPs for a single forward pass of NTK are not as small as FedDST, the sparsity readjustment in FedDST inevitably brings more computation expenses during the training. For the CIFAR-100 dataset, we can observe that the FedDST outperforms other methods in terms of accuracy. Yet, FedDST implements multiple local training epochs (5 epochs in the original paper) to realize the sparsity readjustment, resulting in a heavy computation burden. We further display the experimental results of the NTK pruning given $1$ local training epoch. The accuracy outperforms $5$ epoch FedDST by $0.02\%$, and the computation expense in FLOPs is $5\times$ cheaper than FedDST.
Since the advantages in accuracy brought by the $1$ epoch align with the observations that fewer local steps enhance the global consistency \cite{liu2023enhance}, we conclude that our proposed pruning method enjoys both computation efficiency and better performance.

\begin{figure}[h]  
\vspace{-0.0in}
    \centering
\includegraphics[width=1\linewidth]
    {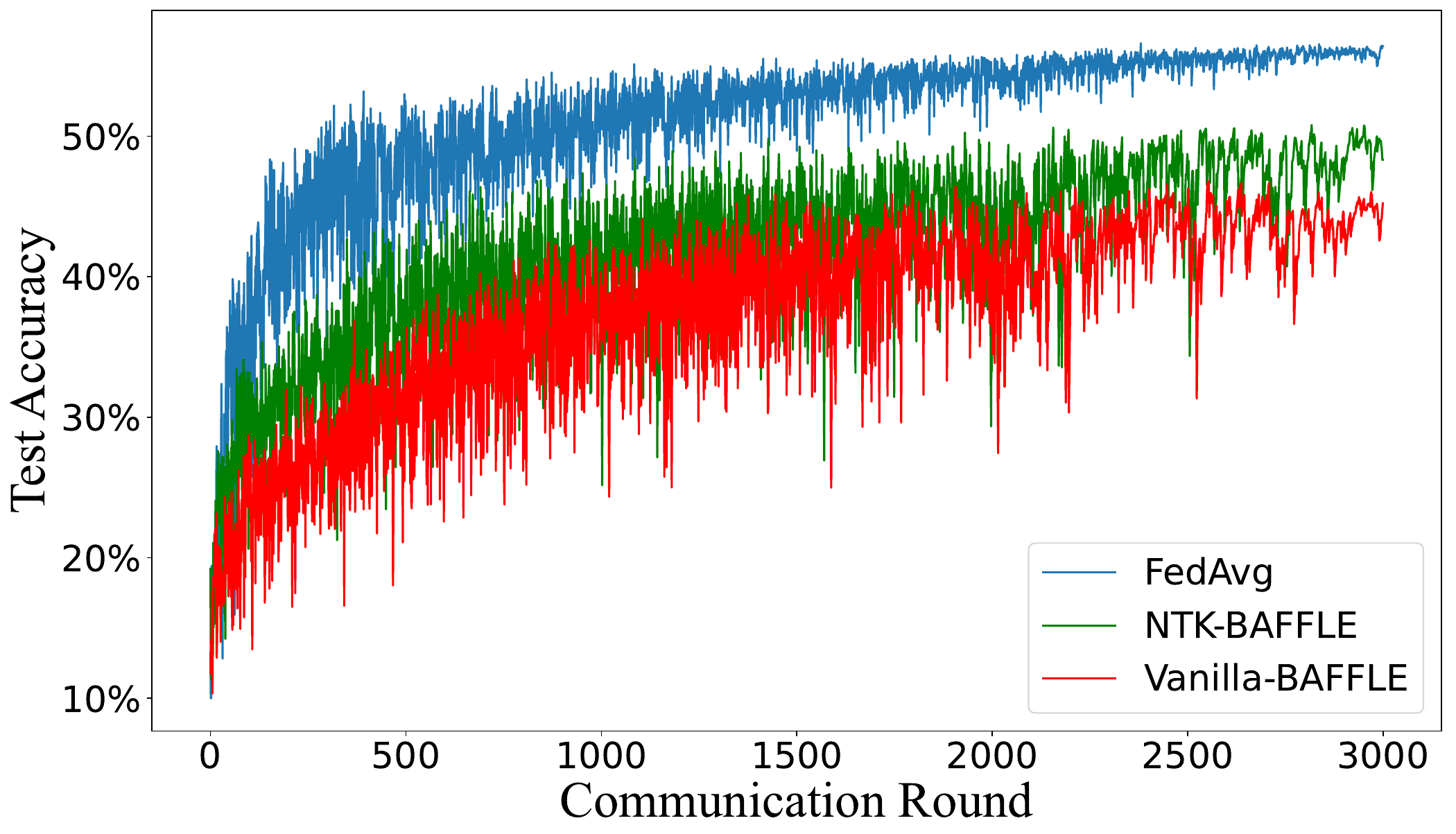}
    \vspace{-0.25in}
    \caption{Test accuracy comparison on LeNet for CIFAR-10. $K$ is set to 200. The communication round is set to 3000.}
    \vspace{-0.05in}
    \label{cifar-10-lenet-fw-200}
\end{figure}

\noindent\textbf{Communication Cost Analysis.}
Assume that the target density is $d$ and the number of parameters of the model is $n$.
PruneFL requires devices
to send full gradients to the server every $\Delta R$ round, resulting in an average upload cost of
$(32d + \frac{32}{\Delta R})n$ bits per device per round, where $\Delta R$ is the interval of sparsity readjustment. Additionally, the maximum upload cost amounts to $(32d + 32)n$ parameters. On the contrary, FedDST has an
average communication cost of $(32d + \frac{1}{\Delta R})n$ bits per device per round before the completion of sparsity readjustment. Once the readjustment is finished, the communication cost decreases to $32dn$. 
In our method, the worst case of communication expense is $32T_pn + 32dn$ bits per device per round.

\noindent\textbf{Data-free Foresight Federated-Pruning Analysis.}
Figure~\ref{cifar-10-non-fw} shows that the result based on randomly generated data is superior to that on the real data.
In addition, the data and label-agnostic properties provide shortcuts to avoid redundant upload and download costs during the pruning step. Using randomly generated data, the 
communication expense during the foresight pruning is reduced to $0$, leading to a communication expense of $32dn$ throughout the training process. As a result, the proposed NTK pruning demonstrates significant advantages in reducing transmission burdens.


\begin{figure*}[h]
\centering
\subfloat[$K=50$]{\includegraphics[width=.33\linewidth]
{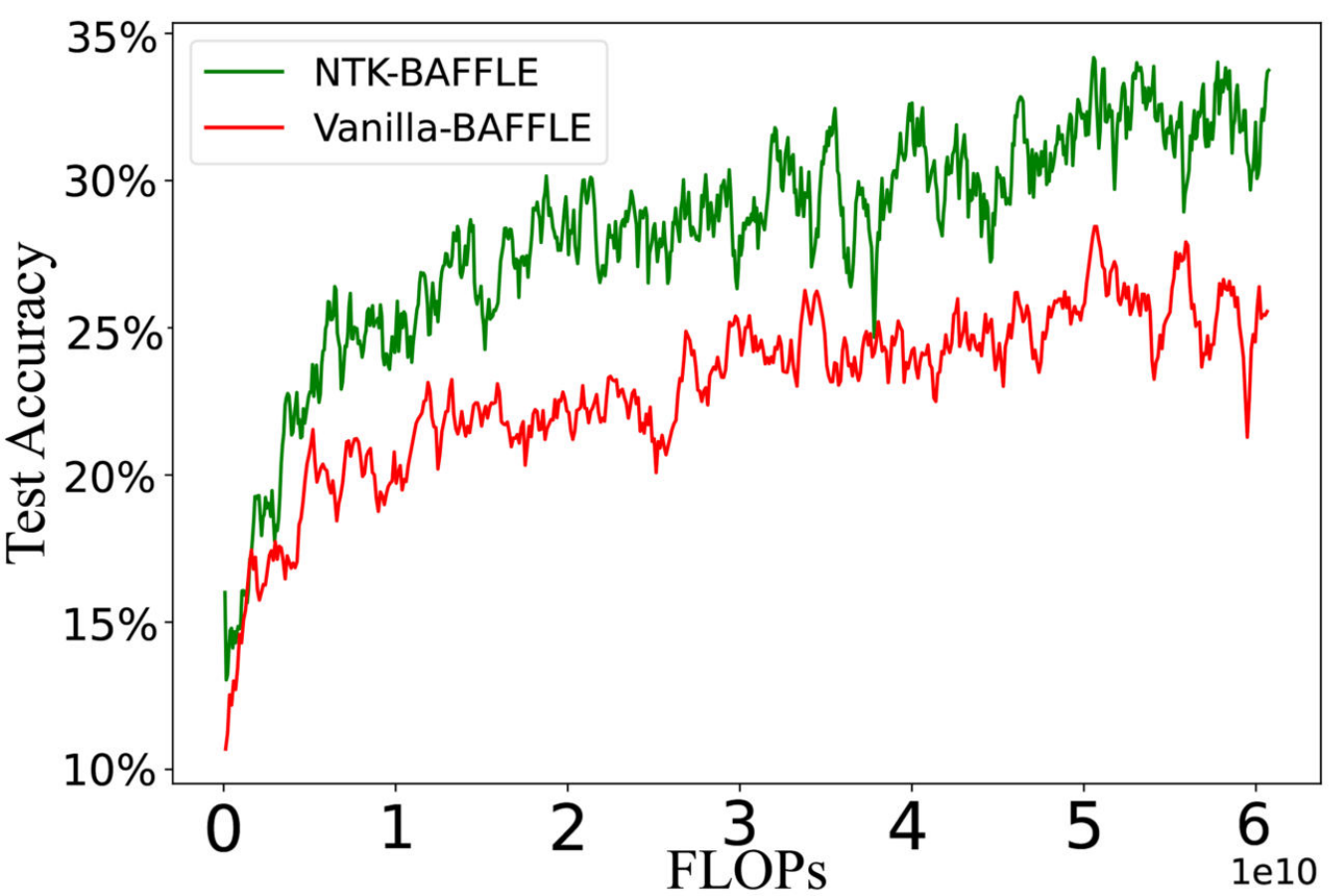}} 
\subfloat[$K=100$]{\includegraphics[width=.33\linewidth]
{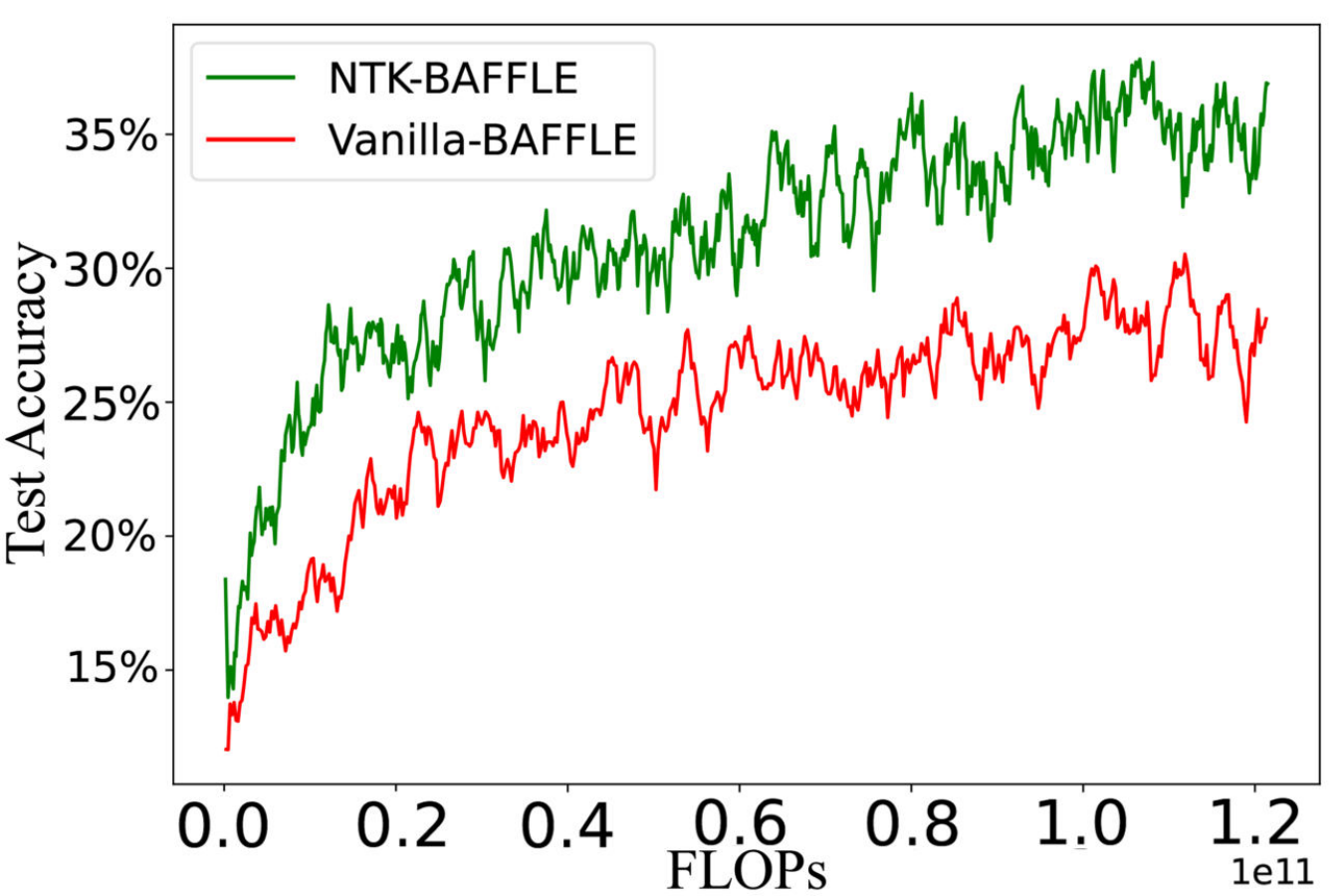}} 
\subfloat[$K=200$]{\includegraphics[width=.33\linewidth]
{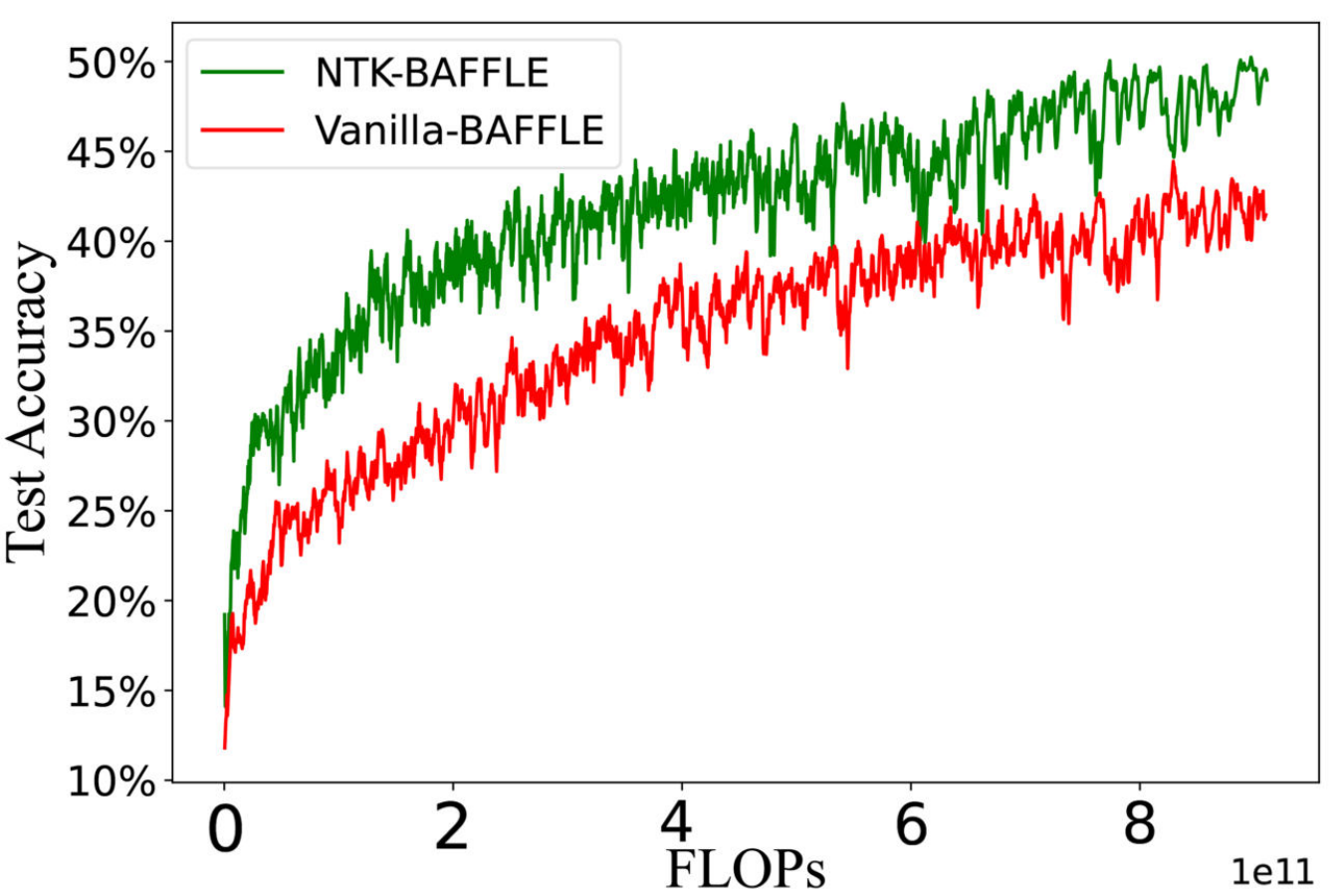}} 
\\ 
\vspace{-0.15in}
\subfloat[$K=50$]{\includegraphics[width=.33\linewidth]
{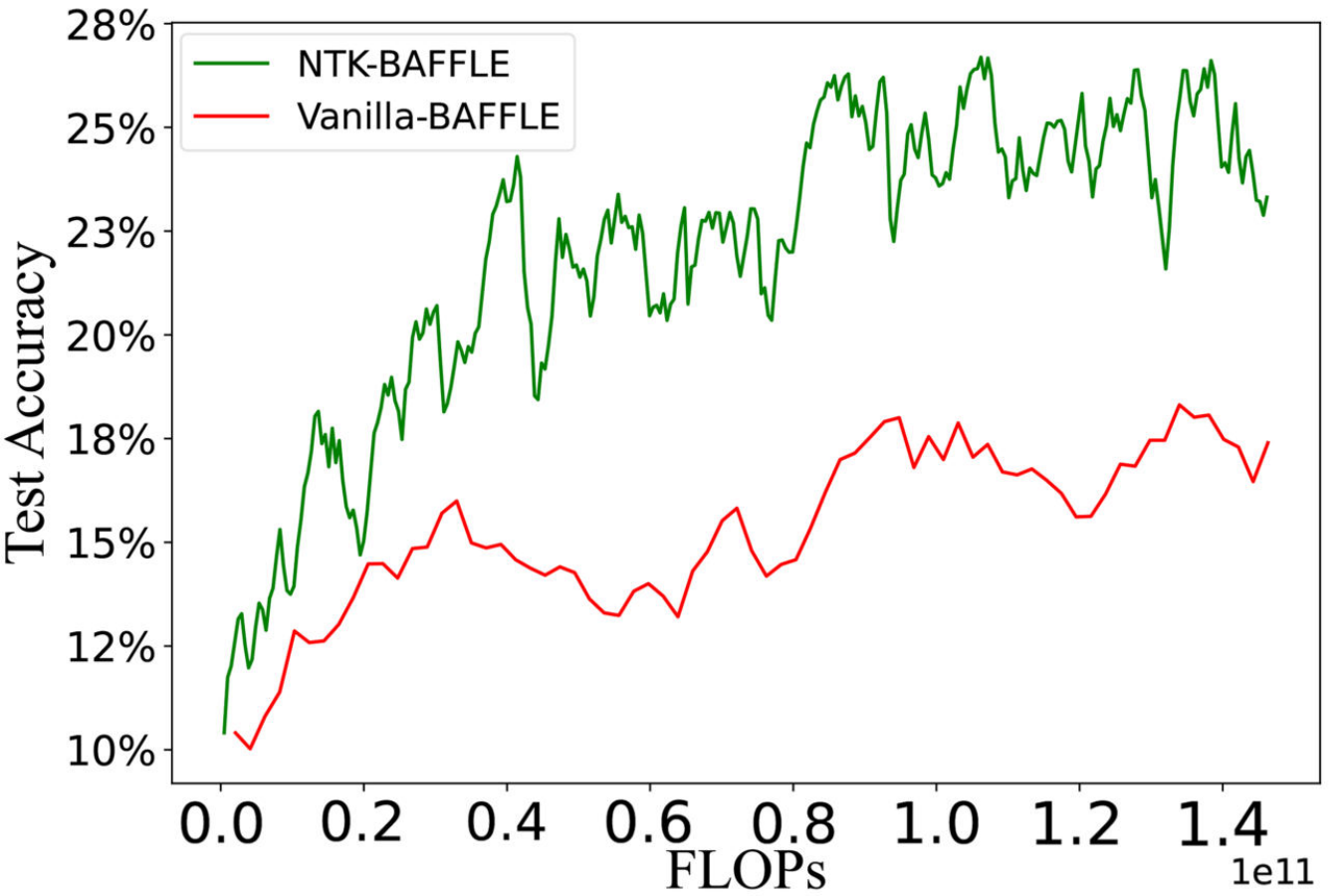}} 
\subfloat[$K=100$]{\includegraphics[width=.33\linewidth]
{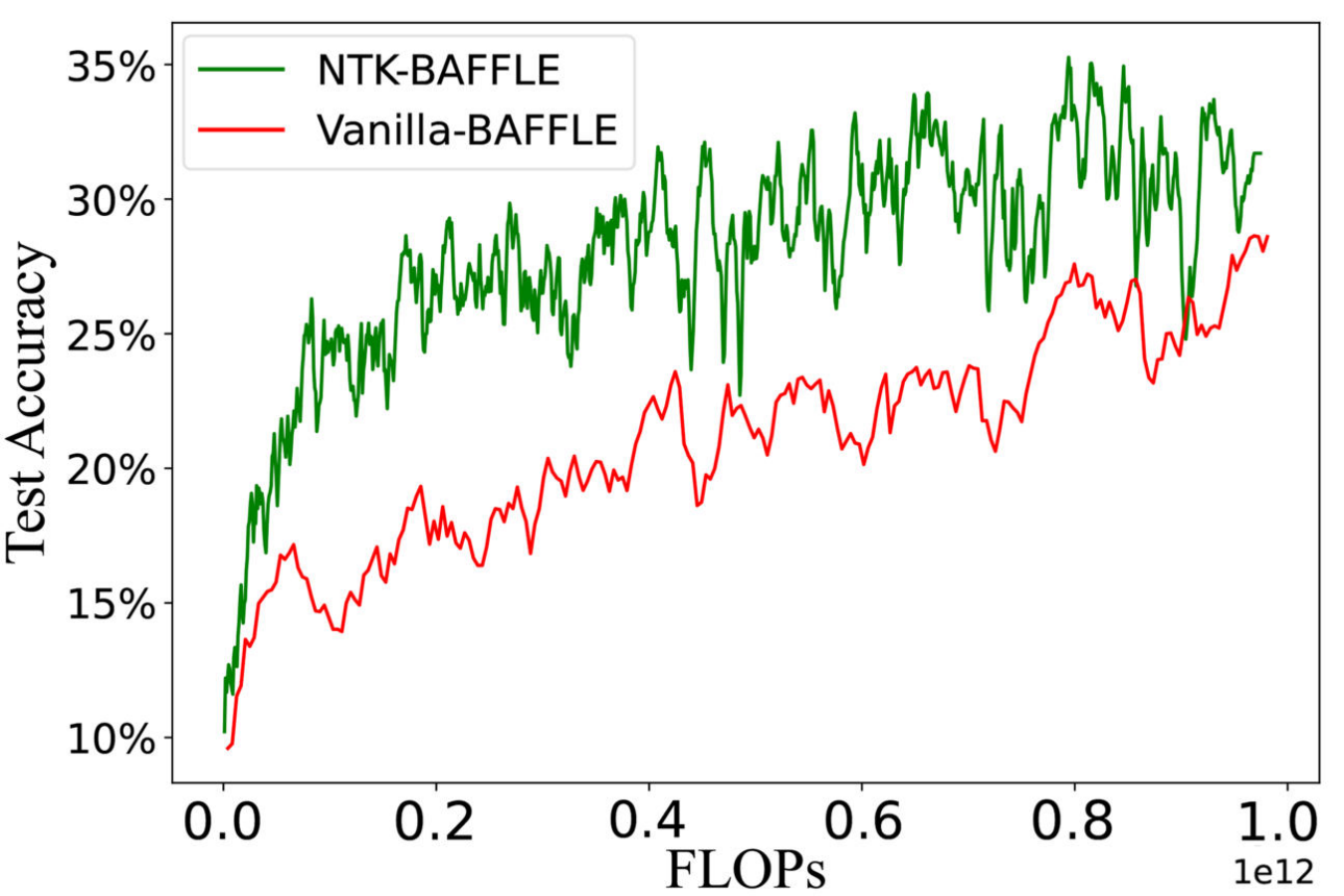}} 
\subfloat[$K=200$]{\includegraphics[width=.33\linewidth]
{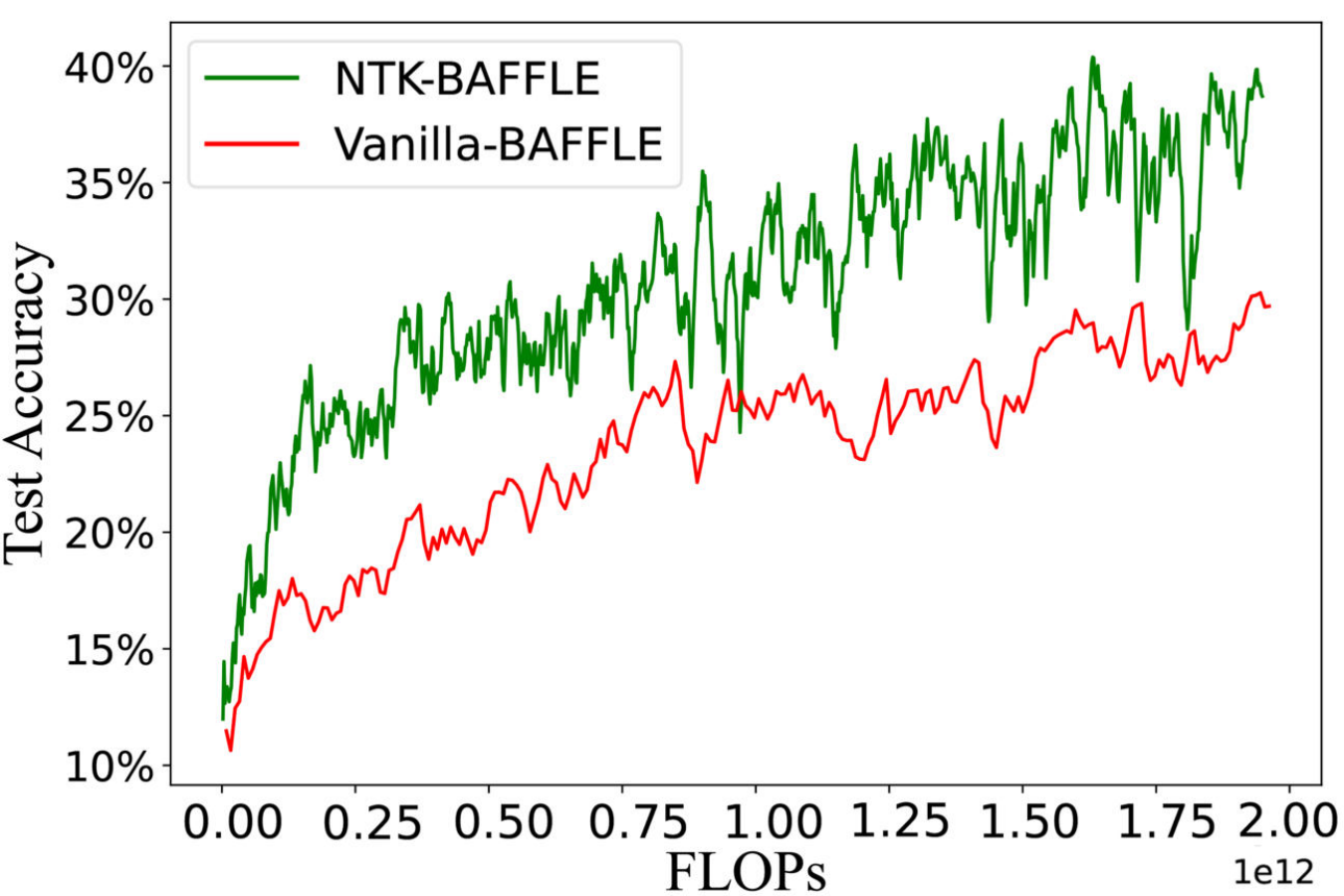}} 
 \vspace{-0.1in}
\caption{Test accuracy for CIFAR-10 with different $K$.
}
\label{cifar-10-fw-ema-all}
\end{figure*}

\subsection{Experimental Results for BAFFLE}

\begin{table}[h]
\centering
\caption{Classification accuracy (\%), consumed FLOPS and peak memory for CIFAR-10/100. }
\small
\begin{tabular}{c|cccc} 
\hline
\multicolumn{1}{c|}{Model} & Method & Max Acc.  & FLOPs & 
\multicolumn{1}{c}{\begin{tabular}[c]{@{}c@{}} Peak\\Memory \end{tabular}}\\ 
\hline
\multirow{4}{*}{LeNet-5} & FedAvg &\multicolumn{1}{c}{$54.07$} &  \multicolumn{1}{c}{$1\textbf{x}$} & $1\textbf{x}$\\
 & FedDST & \multicolumn{1}{c}{$46.43$} & \multicolumn{1}{c}{$0.20\textbf{x}$}  & $1\textbf{x}$\\ 
 & PruneFL &\multicolumn{1}{c}{$52.86$} & \multicolumn{1}{c}{$1\textbf{x}$} & $1\textbf{x}$\\ 
 & NTK-Rand &\multicolumn{1}{c}{$53.65$} & \multicolumn{1}{c}{$0.72\textbf{x}$} & $0.10\textbf{x}$\\ 
\hline
\multirow{5}{*}{ResNet-20} & FedAvg & \multicolumn{1}{c}{$31.69$} & \multicolumn{1}{c}{$1\textbf{x}$} & $1\textbf{x}$\\ 
 & FedDst & \multicolumn{1}{c}{$34.88$} & \multicolumn{1}{c}{$0.43\textbf{x}$} &$1\textbf{x}$\\ 
 & NTK-Rand & \multicolumn{1}{c}{$33.57$} & \multicolumn{1}{c}{$0.43\textbf{x}$} & $1\textbf{x}$\\ 
 & FedAvg(1-ep) & \multicolumn{1}{c}{$34.19$} & \multicolumn{1}{c}{$1\textbf{x}$} &\multicolumn{1}{c}{$1\textbf{x}$}  \\
 & NTK-Rand(1-ep) & \multicolumn{1}{c}{$34.50$} & \multicolumn{1}{c}{$0.43\textbf{x}$} & $0.20\textbf{x}$\\ 
 \hline
\end{tabular}
\vspace{-0.1in}
\label{acc-round}
\end{table}

\noindent\textbf{Classification Accuracy Improvements.} 
Since the performance in Table~\ref{acc-round} and Figure~\ref{cifar-10-non-fw} show that $1$ local epoch and ``NTK-Rand" both maximize the global performance, we set the local training epoch as $1$ and utilize ``NTK-Rand" for the following experiments. 
We denote the vanilla BP-Free method in \cite{does_flzero} as Vanilla-BAFFLE and our method as NTK-BAFFLE. It is worth mentioning that the BP-Free training for FL is currently not competitive with the backpropagation-based FL as the estimated gradients are not as accurate as the true gradients. Therefore, we focus on how many improvements our method can achieve.
Table~\ref{cifar10-fw-table} displays the maximum accuracy comparison between NTK-BAFFLE and Vanilla-BAFFLE on the CIFAR-10 dataset with a sparsity of $90\%$. Note that the number of local training epochs is set to 1. For experiments on the two datasets, the NTK-BAFFLE consistently outperforms the vanilla BAFFLE regarding classification accuracy. For instance, the maximum accuracy is boosted by $4.62\%$, $6.35\%$, and $3.8\%$ on LeNet when setting $K$ to 50, 100, and 200, respectively. Figure~\ref{cifar-10-lenet-fw-200} shows the learning curve against communication rounds given different values of $K$. It's clear that NTK-BAFFLE performs better than the Vanilla-BAFFLE method on various settings of $K$.
Figure~\ref{cifar-10-fw-ema-all} presents the learning curves from the perspective of accuracy versus FLOPs. The first row shows the results on LeNet, and the second row shows that on ResNet-20.

When utilizing ResNet-20, the improvements in the maximum accuracy are $1.00\%$, $1.93\%$, and $1.46\%$, respectively. These phenomena are mainly due to the reason that the estimation over a more complex is not as stable as that over a relatively simple model.
Although the maximum accuracy improvements are not as prominent as that when using ResNet-20, the learning curves against FLOPs from Figure~\ref{cifar-10-fw-ema-all} convincingly demonstrate that NTK-BAFFLE achieves higher accuracy when consuming the same FLOPs.
\begin{table}[b]
\centering
\vspace{-0.1in}
\caption{Maximum accuracy (\%) comparison on CIFAR-10.}
\vspace{-0.1in}
\small
\begin{tabular}{c|c|cll} 
\hline
\multicolumn{2}{c|}{\multirow{2}{*}{Settings}} & \multicolumn{3}{c}{Value of $K$} \\ 
\cline{3-5}
\multicolumn{2}{c|}{} & 50 & 100 & 200 \\ 
\hline
\multirow{3}{*}{LeNet-5} & NTK-BAFFLE &\multicolumn{1}{c}{$\mathbf{39.68}$}  & \multicolumn{1}{c}{$\mathbf{42.70}$} & \multicolumn{1}{c}{$\mathbf{46.39}$}\\ 
\cline{2-5}
 & Vanilla-BAFFLE & \multicolumn{1}{c}{$35.06$} & \multicolumn{1}{c}{$36.35$} & \multicolumn{1}{c}{$42.59$}  \\ 
\cline{2-5}
 & FedAvg & \multicolumn{3}{c}{$\textit{53.20}$} \\ 
\hline
\multicolumn{1}{l|}{\multirow{3}{*}{ResNet-20}} & NTK-BAFFLE & \multicolumn{1}{c}{$\mathbf{35.74}$} &\multicolumn{1}{c}{$\mathbf{38.58}$}  & \multicolumn{1}{c}{$\mathbf{45.27}$}  \\ 
\cline{2-5}
\multicolumn{1}{l|}{} & Vanilla-BAFFLE & \multicolumn{1}{c}{$34.74$} & \multicolumn{1}{c}{$36.65$} & \multicolumn{1}{c}{$43.81$} \\ 
\cline{2-5}
\multicolumn{1}{l|}{} & FedAvg & \multicolumn{3}{c}{$\textit{48.26}$} \\
\hline
\end{tabular}
\label{cifar10-fw-table}
\vspace{-0.2in}
\end{table}

\begin{table}[h]
\centering
\caption{Maximum accuracy (\%) comparison on CIFAR-100.}
\vspace{-0.05in}
\small
\begin{tabular}{c|c|cll} 
\hline
\multicolumn{2}{c|}{\multirow{2}{*}{Settings}} & \multicolumn{3}{c}{Value of K} \\ 
\cline{3-5}
\multicolumn{2}{c|}{} & 50 & 100 & 200 \\ 
\hline
\multirow{3}{*}{LeNet-5} & NTK-BAFFLE &\multicolumn{1}{c}{$\mathbf{5.59}$}  & \multicolumn{1}{c}{$\mathbf{6.80}$} & \multicolumn{1}{c}{$\mathbf{8.35}$}\\ 
\cline{2-5}
 & Vanilla-BAFFLE & \multicolumn{1}{c}{$4.40$} & \multicolumn{1}{c}{$4.94$} & \multicolumn{1}{c}{$5.95$}  \\ 
\cline{2-5}
 & FedAvg & \multicolumn{3}{c}{$\textit{23.52}$} \\ 
\hline
\multicolumn{1}{l|}{\multirow{3}{*}{ResNet-20}} & NTK-BAFFLE & \multicolumn{1}{c}{$\mathbf{12.04}$} & \multicolumn{1}{c}{$\mathbf{14.84}$} & \multicolumn{1}{c}{$\mathbf{17.73}$} \\ 
\cline{2-5}
\multicolumn{1}{l|}{} & Vanilla-BAFFLE & \multicolumn{1}{c}{$11.54$} & \multicolumn{1}{c}{$14.41$}  & \multicolumn{1}{c}{$15.96$} \\ 
\cline{2-5}
\multicolumn{1}{l|}{} & FedAvg & \multicolumn{3}{c}{$\textit{34.19}$} \\
\hline
\end{tabular}
\vspace{-0.1in}
\label{cifar100-fw-table}
\end{table}

\begin{figure*}[ht]
\centering
\subfloat[$K=50$]{\includegraphics[width=.33\linewidth]
{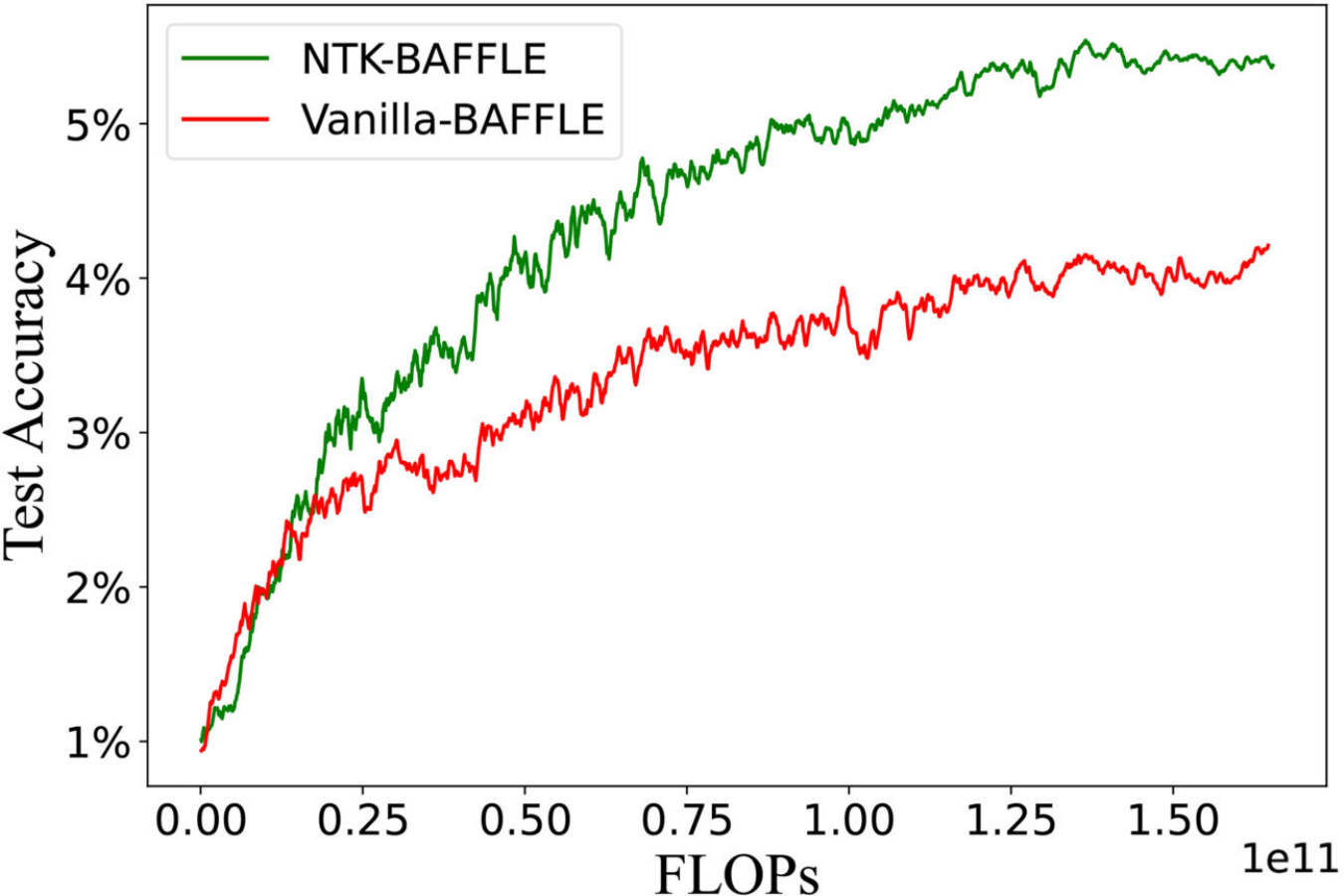}} 
\subfloat[$K=100$]{\includegraphics[width=.33\linewidth]
{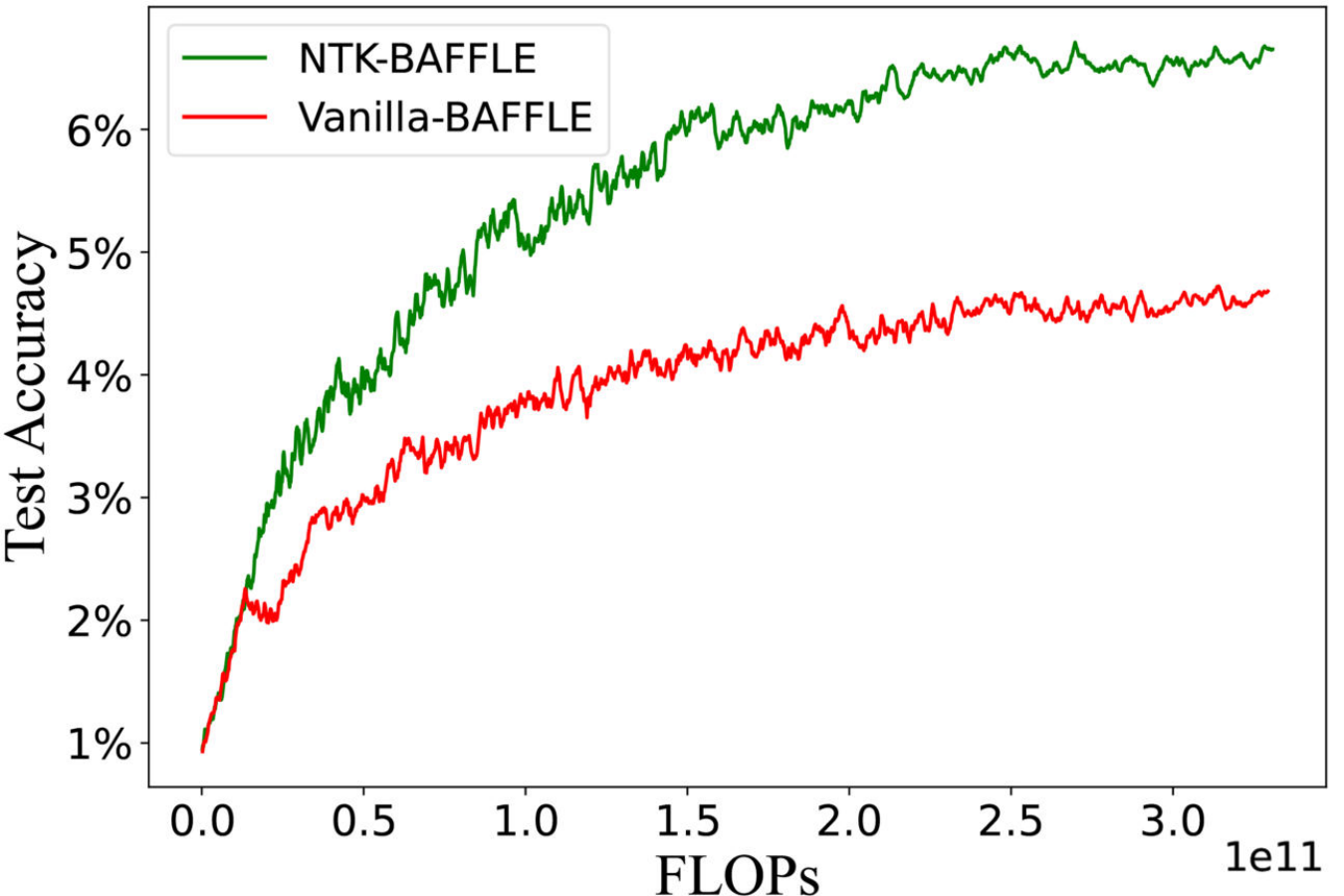}}
\subfloat[$K=200$]{\includegraphics[width=.33\linewidth]
{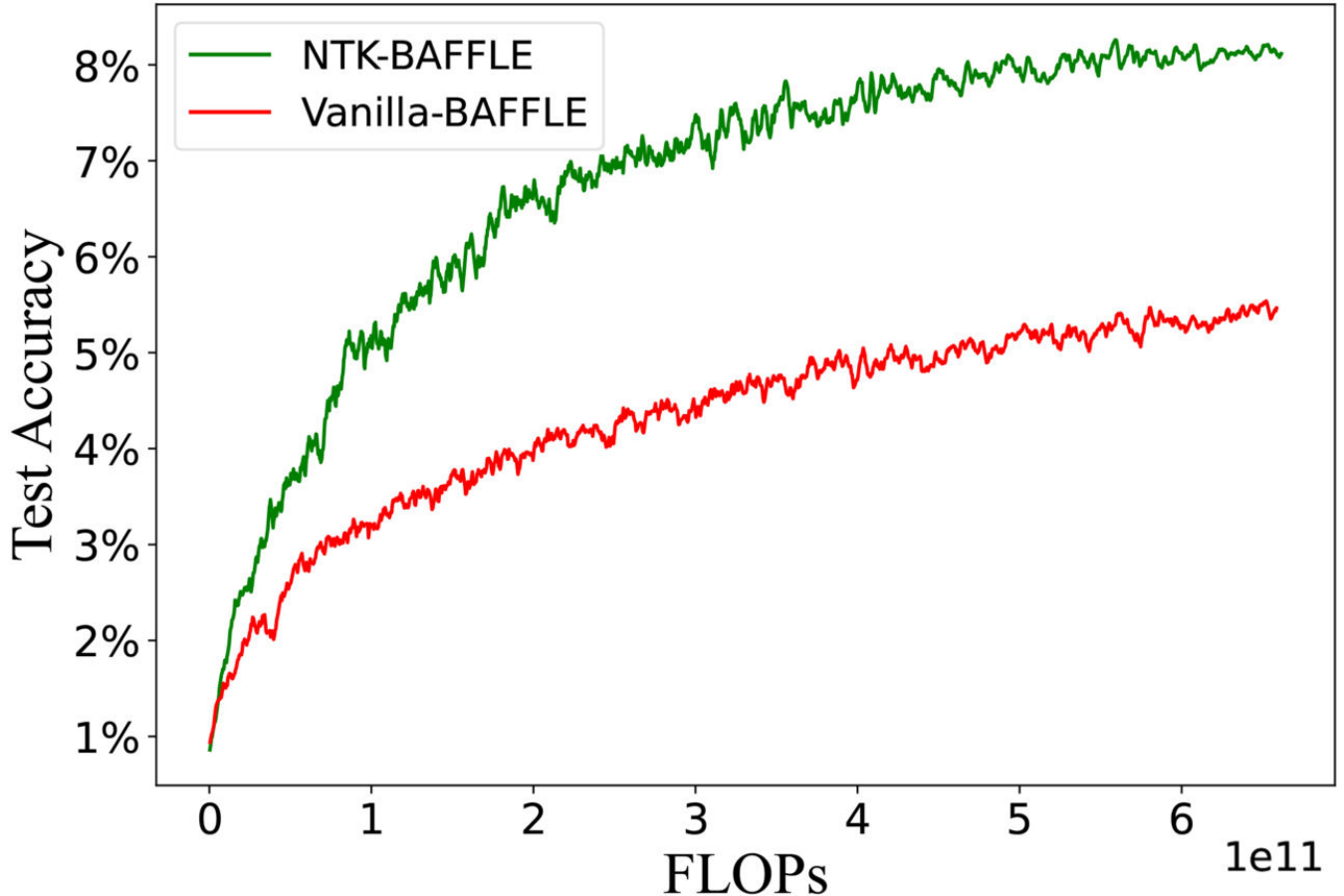}} 
\\
\vspace{-0.15in}
\subfloat[$K=50$]{\includegraphics[width=.33\linewidth]
{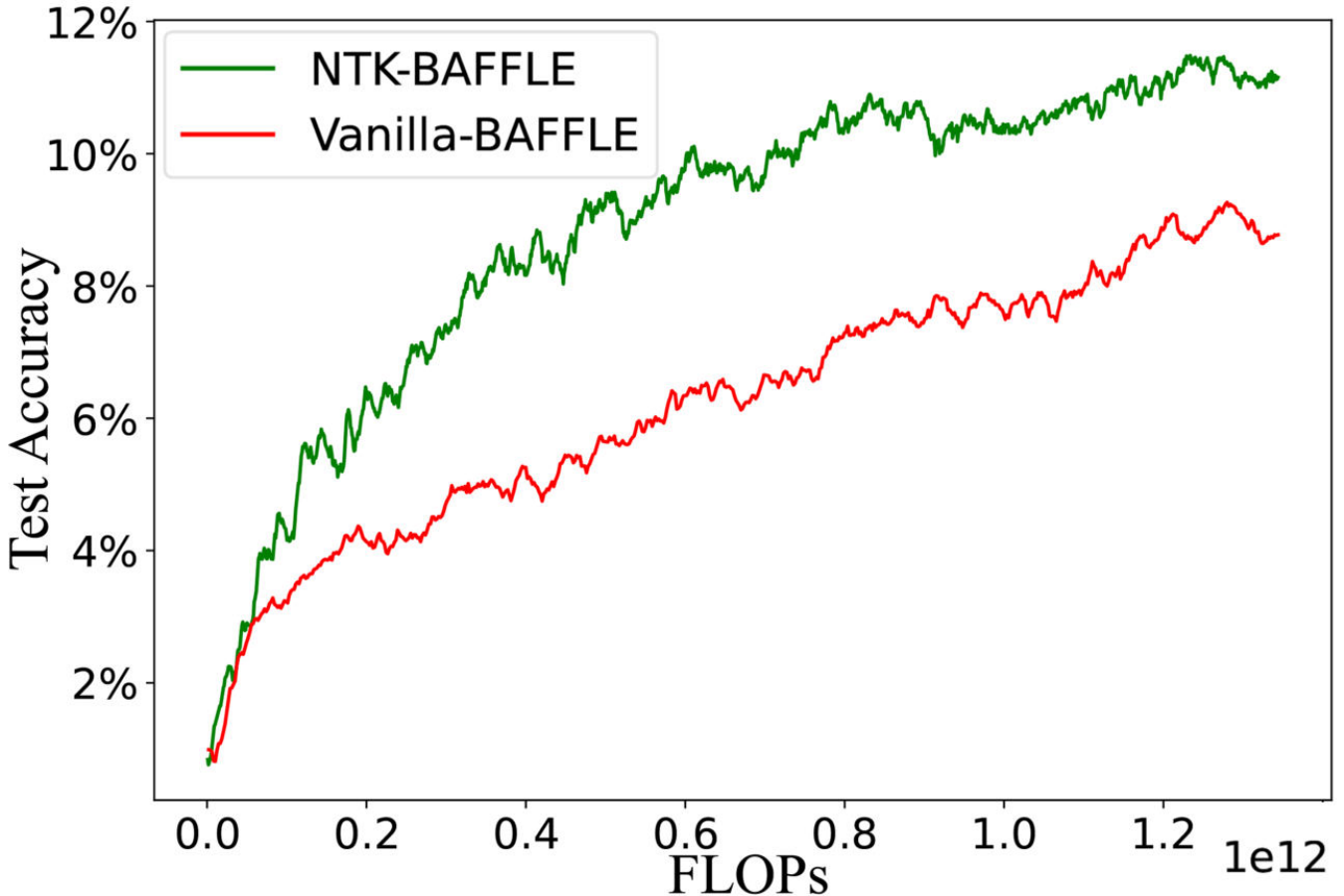}} 
\subfloat[$K=100$]{\includegraphics[width=.33\linewidth]
{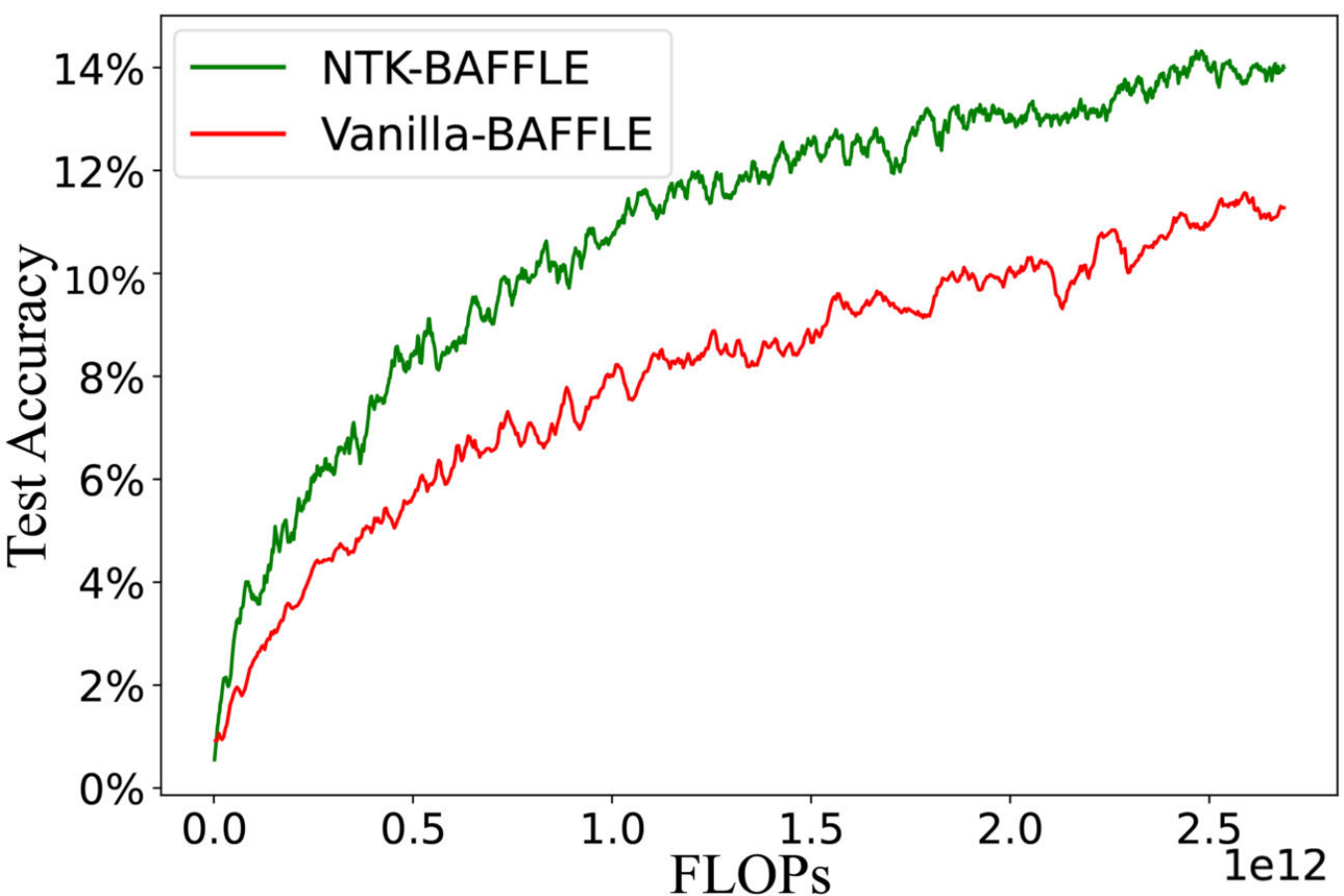}}
\subfloat[$K=200$]{\includegraphics[width=.33\linewidth]
{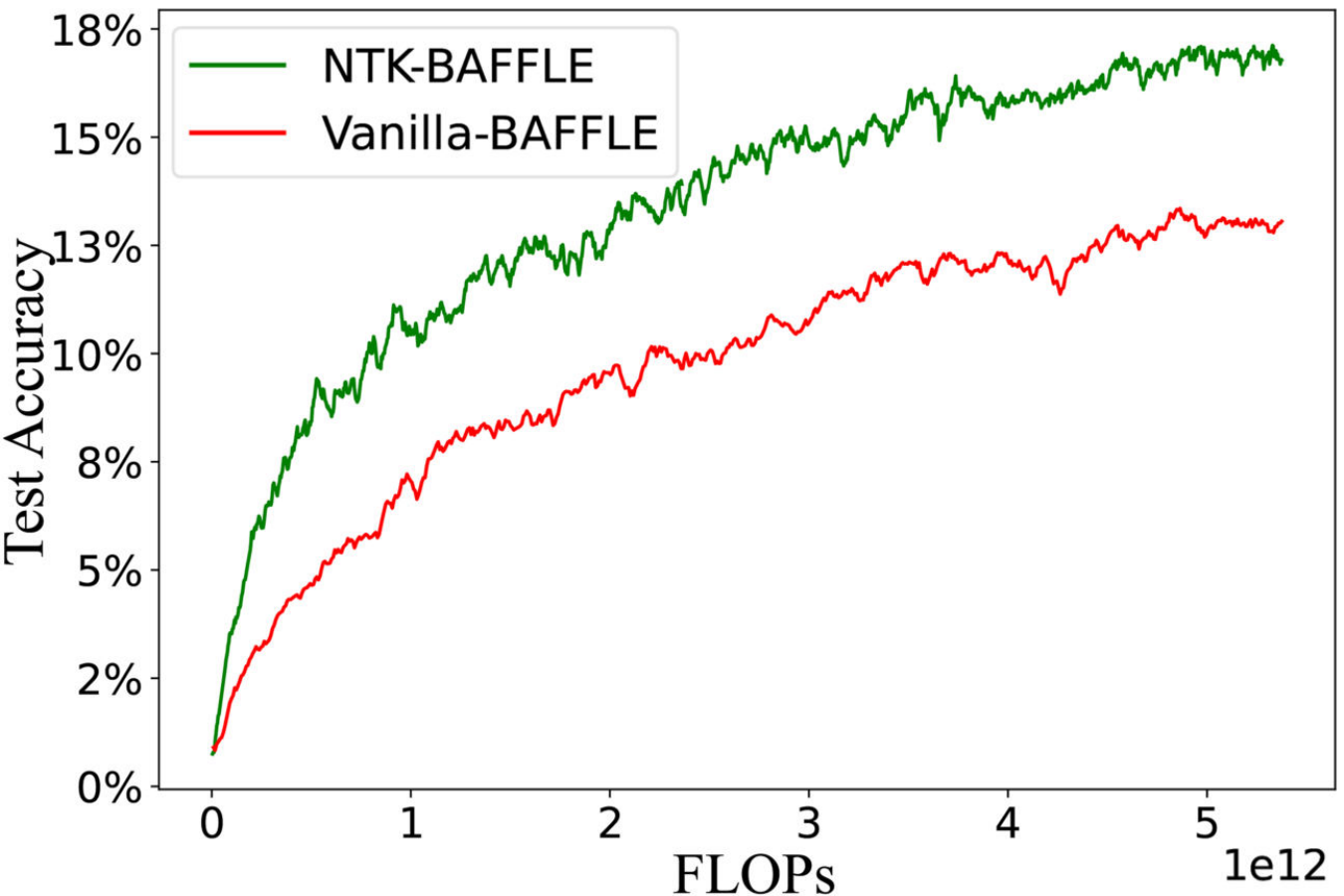}}
 \vspace{-0.1in}
\caption{
Test accuracy for CIFAR-100 with different $K$.}
\label{cifar-100-fw-ema-all}
 \vspace{-0.1in}
\end{figure*}

Table~\ref{cifar100-fw-table} presents the maximum accuracy comparison between NTK-BAFFLE and Vanilla-BAFFLE on the CIFAR-100 dataset. 
Unlike the $90\%$ sparsity setting for the CIFAR-10 dataset, we uniformly set the $80\%$ sparsity level for the CIFAR-100 dataset to ensure the 
Figure~\ref{cifar-100-fw-ema-all} exhibits the learning curves regarding accuracy versus FLOPs.
Same as Figure~\ref{cifar-10-fw-ema-all}, the first row in Figure~\ref{cifar-100-fw-ema-all} presents the results on LeNet, and the second row shows that on ResNet-20.
The maximum accuracy improvements based on the NTK-BAFFLE against Vanilla-BAFFLE are consistently observed through all kinds of settings. For experiments conducted on LeNet, the maximum accuracy is increased by $1.19\%$, $1.86\%$, and $2.4\%$, respectively.
For experiments on ResNet-20, the accuracy is improved by $0.50\%$, $0.43\%$, and $1.77\%$ when setting $K$ to 50, 100, and 200, respectively. From Figure~\ref{cifar-10-fw-ema-all} and Figure~\ref{cifar-100-fw-ema-all}, we can notice that our proposed pruning method consistently enhances the performance of the original BAFFLE method by achieving higher accuracy while consuming lower FLOPs. 

\begin{table}[h]
\centering
\vspace{-0.1in}
\caption{Platform configuration.}
\vspace{-0.05in}
\small
\begin{tabular}{l|c|c|c} 
\hline
\multicolumn{1}{c|}{Type} & Device            & Configuration        & \multicolumn{1}{l}{Number}  \\ 
\hline
\multirow{2}{*}{Device}   & Jetson Nano       & 128-core Maxwell GPU & 7                           \\
                          & Jetson Xavier AGX & 512-core NVIDIA GPU  & 3                           \\ 
\hline
Server                    & Workstation       & NVIDIA RTX 3080 GPU  & 1                           \\
\hline
\end{tabular}

\label{platform}
\vspace{-0.05in}
\end{table}

\begin{figure}[b]
\vspace{-0.35in}
\centering
\subfloat[Devices of our real test-bed]{\includegraphics[width=0.46\linewidth]{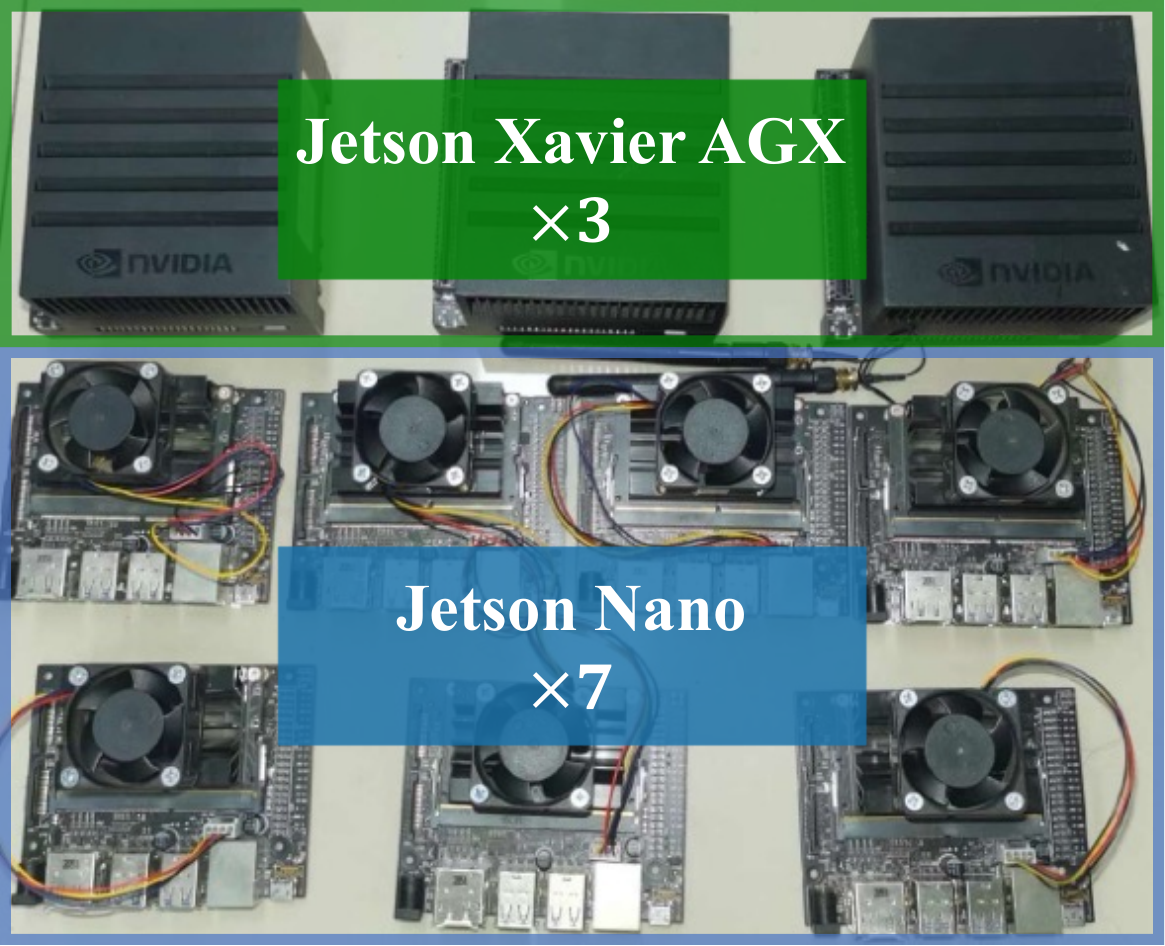}}  
\subfloat[Learning curves]{\includegraphics[width=0.52\linewidth]{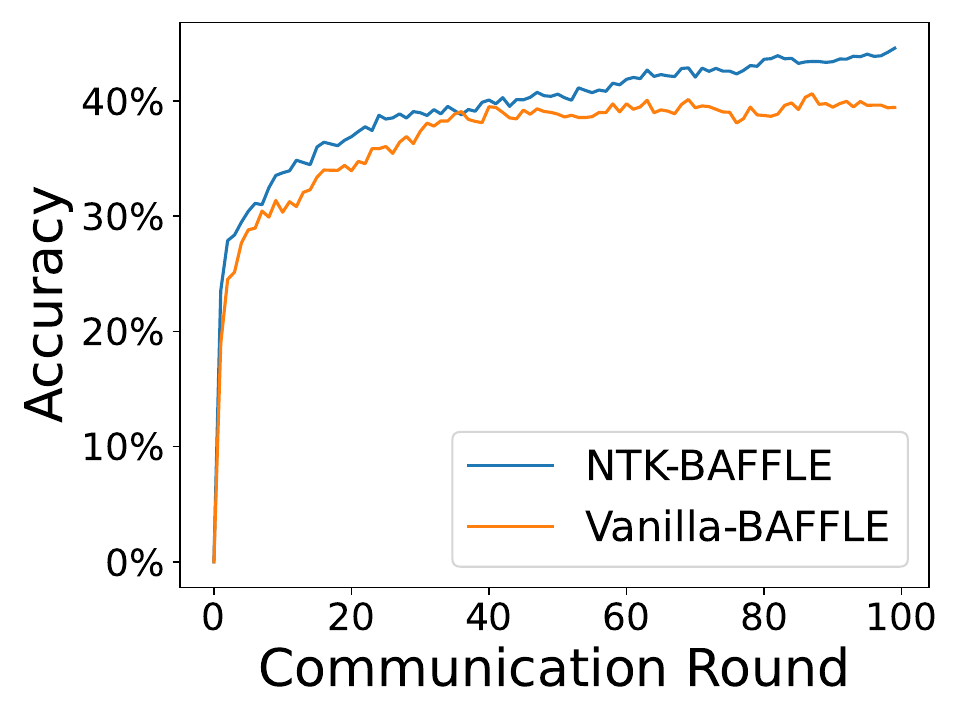}} 
\vspace{-0.05in}
\caption{Learning performance on the real test-bed.}
\label{platform_acc_round}
\vspace{-0.2in}
\end{figure}

\noindent\textbf{Real Test-bed Results.}
We conducted real test-bed experiments on CIFAR-10 under the $\beta=0.1$ heterogeneity. The model we utilized is the LeNet-5. The training hyperparameters stay the same and the number of Monte Carlo steps $K$ is set to $50$.
The platform configuration is listed in Table~\ref{platform}. It is composed of $10$ device devices and one cloud server. $2$ devices are randomly selected in each training round.
Figure \ref{platform_acc_round} shows the device devices and the comparison of learning curves. We can find that our method consistently beats the Vanilla-BAFFLE in classification accuracy.

\noindent\textbf{Memory Analysis.}
The Vanilla-BAFFLE and NTK-BAFFLE offer an effective solution for minimizing memory usage on edge devices, combining the benefits of static and dynamic memory efficiency. When it comes to running backpropagation on deep networks using an auto-differential framework, additional static memory is needed. 
This places a significant strain on memory-constrained devices. Moreover, backpropagation relies on storing intermediate activations, leading to substantial requirements for dynamic memory.
In contrast, the additional memory to store the perturbation in each forward pass and estimate the local gradient for the BAFFLE-based method is $O(n)$. To maximize the benefits of the BP-Free method, we perform layer-by-layer computations, effectively partitioning the forward computation graph into smaller segments.
Table~\ref{memory} shows the peak GPU memory usage (MB) during the forward pass.
For LeNet, the peak memory usage of Vanilla-BAFFLE and NTK-BAFFLE is $10.26\%$ of the BP operation. For ResNet-20, the peak memory of BAFLLE-related methods is $19.74\%$ of the BP process.
Since NTK-BAFFLE implements sparse training, additional memory is needed to store the binary mask. However, modern frameworks (e.g., \cite{XNNPACK}) are capable of efficiently storing and processing sparse matrices. 
In addition, the binary mask can be efficiently stored with a $1$-bit datatype.
As a result, the extra memory usage associated with NTK-BAFFLE is negligible.

\begin{table}[ht]
\centering
\caption{The peak GPU memory cost (in MB, including feature maps and parameters) of vanilla backpropagation (BP), Vanilla-BAFFLE, and NTK-BAFFLE.}
\small
\begin{tabular}{c|cll} 
\hline
Model & BP & Vanilla-BAFFLE & NTK-BAFFLE \\ 
\hline
LeNet & \multicolumn{1}{c}{$1617$}  & \multicolumn{1}{c}{$166$} & \multicolumn{1}{c}{$\mathbf{166}$} 
\\
ResNet-20 & \multicolumn{1}{c}{$1671$} & \multicolumn{1}{c}{$330$} & \multicolumn{1}{c}{$\mathbf{330}$} \\
\hline
\end{tabular}

\label{memory}
\end{table}

%
The additional memory required for storing the perturbation in each forward pass and estimating the local gradient for the BAFFLE-based method is $O(n)$.
%
%
\noindent It is worth noting that modern frameworks (e.g., \cite{XNNPACK}) can efficiently store and process sparse matrices, making the memory needed to store the binary mask associated with NTK-BAFFLE negligible.
We perform layer-by-layer computations, partitioning the forward computation graph into smaller segments and calculating the peak GPU memory usage during the forward pass when the input is $32$ RGB images with a size of $32\times 32$  from the CIFAR-10 dataset. 
In LeNet and ResNet-20, the peak memory usage of Vanilla-BAFFLE is significantly smaller compared to backpropagation-based methods, with values of $166$ MB vs. $1617$ MB and $330$ MB vs. $1671$ MB, respectively.
Additionally, the proposed NTK-BAFFLE has fewer parameters. When considering an $80\%$ pruning rate, the peak memory usage of NTK-BAFFLE due to parameters is only $7.44\%$ (LeNet) and $3.88\%$ (ResNet-20) of the usage in Vanilla-BAFFLE, respectively.

\noindent\textbf{Communication Overhead Analysis.}
In the BP-based FL system, the communication cost composed by uploading and downloading is twice the number of model parameters.
In Vanilla-BAFFLE and NTK-BAFFLE, each device only uploads a $K$-dimensional vector to the server for aggregation. Since $K$ is significantly
less than the parameter amounts $n$, both the Vanilla-BAFFLE and NTK-BAFFLE greatly reduce the cost during the upload compared to the conventional backpropagation-based
FL. For download, NTK-BAFFLE requires less data transmission due to the sparsity property compared to Vanilla-BAFFLE. Overall, NTK-BAFFLE has the lowest communication costs.


\section{Conclusion}
Although Federated Learning (FL) is becoming increasingly popular in AIoT design, 
it  greatly suffers from the problem of low inference performance when 
dealing with memory-constrained AIoT devices.
To address this issue, in this paper 
we proposed a memory-efficient federated foresight pruning method based on the NTK theory. 
Since our proposed foresight-pruning method can seamlessly integrate into BP-Free training frameworks, it can significantly reduce the memory footprint during FL training. Meanwhile, by combining the proposed foresight pruning with the finite difference method used in BP-Free training, the performance of the vanilla BP-Free method is optimized, thus significantly reducing the overall FLOPs of FL. 
Note that, due to the introduction of  a data-free approach to diminishing the error of using local NTK matrices to approximate the federated NTK matrix, our approach is 
resilient to various extreme data heterogeneity scenarios. 
%
%
%
Comprehensive experimental results obtained from simulation- and real test-bed-based platforms with various DNN models 
demonstrate the effectiveness of our method from the perspectives of memory usage and computation burden.

\clearpage
\bibliographystyle{ACM-Reference-Format}
\bibliography{acmart}

\end{document}